\documentclass[letterpaper, 10 pt, conference]{ieeeconf}  

\IEEEoverridecommandlockouts                              

\usepackage{amsmath} 
\usepackage{graphicx} 
\usepackage{float}
\usepackage{subcaption}
\usepackage{listings}
\usepackage{array}
\usepackage[percent]{overpic}

\title{\LARGE \bf
RobotDesignGPT: Automated Robot Design Synthesis \\
using Vision Language Models
}

\author{
Nitish Sontakke$^{1}$ K. Niranjan Kumar$^{1}$ Sehoon Ha$^{1}$
\thanks{$^{1}$School of Interactive Computing, Georgia Institute of Technology, Atlanta, GA, 30308, USA. {\tt \small nitishsontakke@gatech.edu, niranjankumar@gatech.edu, sehoonha@gatech.edu}}%
}

\usepackage{xcolor}
\newcommand{\cmt}[1]{}

\long\def\ignorethis#1{}

\newcommand{\etal}{{\em{et~al.}\ }}

\begin{document}

\maketitle
\thispagestyle{empty}
\pagestyle{empty}

\begin{abstract}

Robot design is a nontrivial process that involves careful consideration of multiple criteria, including user specifications, kinematic structures, and visual appearance. Therefore, the design process often relies heavily on domain expertise and significant human effort. 
The majority of current methods are rule-based, requiring the specification of a grammar or a set of primitive components and modules that can be composed to create a design. We propose a novel automated robot design framework, \textbf{RobotDesignGPT}, that leverages the general knowledge and reasoning capabilities of large pre-trained vision-language models to automate the robot design synthesis process. Our framework synthesizes an initial robot design from a simple user prompt and a reference image. Our novel visual feedback approach allows us to greatly improve the design quality and reduce unnecessary manual feedback. We demonstrate that our framework can design visually appealing and kinematically valid robots inspired by nature, ranging from legged animals to flying creatures. We justify the proposed framework by conducting an ablation study and a user study.

\end{abstract}
\section{INTRODUCTION}

Robot design is a fundamental task in robotics, which significantly influences the final performance. This task can be roughly categorized into two subproblems: design optimization and design synthesis. In design optimization, the algorithm receives an initial design as input and then tries to optimize the continuous design parameters, such as link lengths and body dimensions, using gradient-based optimization methods. However, the performance of such an optimization is often heavily dependent on the initial design. This brings us to robot design synthesis, which aims to generate effective designs from scratch based on the user description. Despite its tremendous potential, synthesizing robot designs is a challenging problem because it requires a deep understanding of both mechanical design and motion control. These challenges limit the choice of algorithms to relatively simple approaches, such as evolutionary algorithms~\cite{gupta2021embodied, zonghao2022evorobogami}.

\begin{figure}
    \centering
    \includegraphics[scale=0.325]{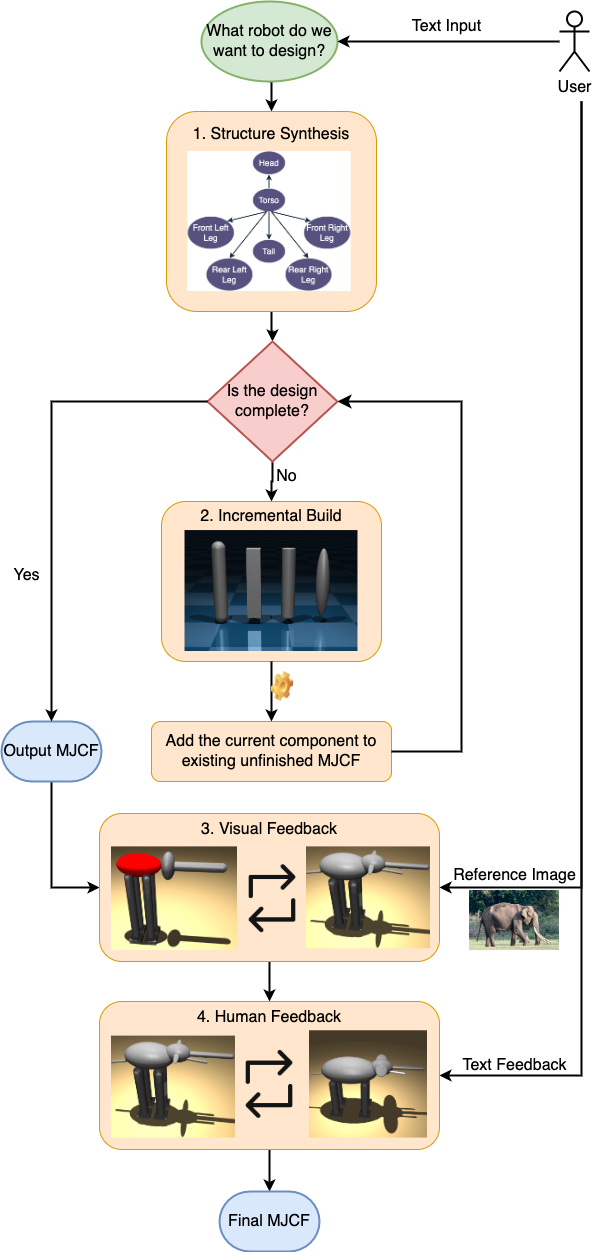}
    \caption{Flowchart depicting an overview of our method.
    }
    \label{fig:flowchart}
\end{figure}

In this work, we leverage Vision Language Models (VLMs) to advance the state-of-the-art in robot design synthesis. We draw inspiration from recent advances in various robotics domains such as manipulation~\cite{stone2023open, huang2023voxposer, duan2024manipulate} and navigation~\cite{shah2023lm, zhou2024navgpt, yokoyama2024vlfm} that use VLMs for open-ended reasoning. VLMs are well-suited for robot design synthesis and provide a natural language interface for the user to describe the requirements. Our work focuses on two specific features of VLMs that have special significance for robot synthesis. First, their baked-in general knowledge enables them to understand the anatomy of different species and in turn reason about composing complex structures with multiple simple shapes. Second, VLMs can directly translate this knowledge into commonly used robot description formats, such as URDF~\cite{urdf_wiki} and MJCF~\cite{todorov2012mujoco}. While promising, a naive application of VLMs using few-shot prompts is insufficient for synthesizing valid robot designs. This is because robot synthesis involves long contexts and spatial reasoning which often lead to hallucinations from the VLM that deviate from the space of physically valid designs. 

We propose a framework for synthesizing kinematically valid robot designs based on simple user inputs. Our approach takes as input a text label (such as ``rhinoceros'' or ``ladybug'') and an optional reference image and first generates a structural design by querying the VLM. It then iterates through the structural elements using our RobotDesignGPT API to synthesize joints and bodies. To ensure connectivity and kinematic validity, we provide compact APIs specifically designed for VLMs. We develop a visual-based automated feedback mechanism that interprets the generated designs, compares them with the user-provided text descriptions and image inputs, and automatically improves on the designs iteratively. Finally, the user has the option to provide stylistic preferences to the VLM to allow for further customization.

We demonstrate that the proposed framework generates a diverse range of robot designs, such crab, dragonfly, rabbit, penguin, ostrich, and more, using just $2.15$ iterations of human feedback on average. The generated designs then can be further integrated with motion planning modules to generate plausible robot motions. To evaluate the quality of our results with independent viewers, we conducted a user study with 44 respondents who were asked to rate the generated robot designs in terms of how closely they resemble the reference creature, achieving an average user satisfaction score of $2.93$ out of $5$. We also conducted an ablation study to show that automated visual feedback is crucial for generating convincing robot designs. To the best of our knowledge, this is the first framework capable of synthesizing kinematically valid articulated rigid body designs from user inputs without relying on manually defined rules or templates.

Our technical contributions are as follows.
\begin{enumerate}
    \item We propose a VLM-based design synthesis framework capable of generating diverse kinematic robot designs, from legged robots to flying creatures,
    \item We further enhance design quality by incorporating automated visual feedback from a reference image and limited human feedback,
    \item We demonstrate that the generated designs are kinematically valid and capable of producing compelling motions using off-the-shelf motion planners.
\end{enumerate}




\section{RELATED WORK}

\subsection{Robot Co-Design via Optimization}
The performance of a robot on a given task depends on its motion controller, which is in turn tightly coupled with the robot's design. Automated robot design aims to maximize the performance with respect to the given task by simultaneously optimizing the robot's design parameters and its motion controller. Current state-of-the-art methods include evolutionary algorithms~\cite{gupta2021embodied, zonghao2022evorobogami}, guided optimization with user-specified motion trajectories~\cite{ha2017joint, ha2018tro, ha2018computational}, grammar-guided approaches~\cite{zhao2020robogrammar, zhao2022automatic, hu2023glso}, and Bayesian optimization~\cite{bjelonic2023learning, pan2021emergent}. More recently, the research community has also observed an increase in learning-based approaches, including deep reinforcement learning~\cite{schaff2022soft, li2024reinforcement, he2024morph}, meta reinforcement learning~\cite{belmonte2022meta}, hierarchical reinforcement learning~\cite{sun2023co}, transformer-based methods~\cite{yu2023multi}, and diffusion-based methods~\cite{wang2024diffusebot, xu2024dynamics}. The use of natural language as an input modality, such as in Text2Robot~\cite{ringel2024text2robot}, is also being actively explored.



\subsection{Large Language and Vision Models in Robotics}
Large Language Models (LLMs) and Vision Language Models (VLMs) have emerged as innovative solutions in numerous research domains. These models have also been effectively leveraged for various robotic use cases, such as manipulation~\cite{stone2023open, huang2023voxposer, duan2024manipulate}, navigation~\cite{shah2023lm, zhou2024navgpt, yokoyama2024vlfm}, mobile manipulation~\cite{yenamandra2023homerobot, wu2024helpful, nasiriany2024pivot}, and locomotion~\cite{wang2023prompt, tang2023saytap}, by offering baked-in general knowledge coupled with commonsense and chain-of-thought reasoning capabilities. This ability to reason over long horizons makes them particularly effective as task and motion planners. To this end, we have seen LLMs and VLMs being successfully utilized in task and motion planning (TAMP)~\cite{singh2023progprompt, ding2023task, sun2024prompt}.
In addition, LLMs and VLMs have also been integrated into various parts of the developmental pipeline itself, including reward engineering \cite{ma2023eureka, yu2023language, xietext2reward}, code generation 
\cite{liang2023code, arenas2024prompt}, motion generation~\cite{jiang2024harmon}, curriculum generation~\cite{ryu2024curricullm}, domain randomization~\cite{ma2024dreureka},
task decomposition~\cite{ahn2022can, katara2024gen2sim}, real-to-sim~\cite{katara2024gen2sim}, and a combination of several of the above~\cite{xu2023creative}.

\subsection{Large Language Model-based Design Workflows}
The recent advances of LLMs and VLMs also open new possibilities for robot design frameworks and algorithms, allowing them to incorporate generic human knowledge. Stella \etal~\cite{stella2023can} utilize an LLM to design a robotic gripper in collaboration with a human for the tomato harvesting task. Makatura \etal~\cite{makatura2023can, makatura2024can} demonstrate how LLMs can be incorporated in each phase of the computational design process and demonstrate impressive results by building a cabinet and fabricating a functional quadcopter. LLMs and VLMs have also been shown to be effective for 3D modeling tasks \cite{yuan20243d, sun20233d, le2024articulate}. Ma~\etal \cite{ma2024exploration} leverage them to design soft modular robots. RoboMorph~\cite{qiu2024robomorph} combines several of the techniques mentioned earlier to create modular robots as it employs graph grammar as part of the LLM's input prompt to encode structural constraints, evolutionary algorithms to improve the designs, and reinforcement learning to learn motion controllers.

Inspired by these recent advancements, we propose an end-to-end method that uses a VLM backbone to design a robot model from user-provided texts and images. In contrast to other methods, our approach uses a combination of text and image inputs to automatically generate articulated robots with arbitrary geometries which can be actuated. Furthermore, we focus on robot design synthesis, the output of which can then be used for design optimization. This enables users to combine our method with several of the existing design optimization methods.

\section{RobotDesignGPT}
We propose a novel framework, \textbf{RobotDesignGPT}, which allows users to synthesize robot description files from descriptive text and image inputs automatically. We adopt the MuJoCo Modeling XML File (MJCF)\cite{todorov2012mujoco} format in our implementation.
Our framework does not impose any specific assumptions, supporting the synthesis of a wide range of robots, from legged robots to flying creatures. Furthermore, our framework does not require any expert knowledge such as an understanding of robot descriptions or the crafting of objective or reward functions.



Figure~\ref{fig:flowchart} provides a visual summary of our method. In this section, we provide a detailed description of each of the four stages.
\subsection{\textbf{Structure Synthesis}}
For the given language input, such as ``penguin'' or ``kangaroo'', our framework first prompts a VLM to synthesize the robot's kinematic tree. Specifically, we instruct the VLM to generate a hierarchy of body nodes represented as a tree structure to capture the parent-child relationships. However, multiple valid structures can exist for the same input. For example, a ``spider'' robot should have eight legs, but each leg can consist of two, three, or even more links. While a greater number of links can improve visual fidelity, it also exponentially increases the complexity of the search space and often leads to invalid or inconsistent robot designs. To address this issue, we include a preference for such design choices as part of the text prompt. To constrain the space of possible designs we additionally provide prompts limiting the total number of components, the maximum number of links per component, symmetry requirements, and other factors.

\subsection{\textbf{Incremental Build}} \label{sec:build}
In the next step, our framework iterates over the body nodes and generates the necessary components: body dimensions, geometric primitives, joint location, and joint type. Roughly, this procedure can be viewed as populating the properties of ``body,'' ``geom,'' ``joint,'' and ``actuator'' tags in an MJCF.

\noindent \textbf{Joint Location.}
For each body node, except for the root, the framework must determine where the child link begins by specifying a joint location. However, this is not a straightforward decision as determining an exact joint location requires a spatial understanding of the desired robot morphology and the geometric primitives of the parent node. In our experience, directly asking a VLM to determine a joint position often results in an invalid structure with either noticeable gaps or penetrations. Instead, we prompt the VLM to determine the direction of growth $\mathbf{d}_{\text{grow}}$ from the parent's center $\mathbf{c}_{\text{parent}}$ and compute the intersection between the ray and the primitive. We compute the exact location by solving the following optimization:
\begin{equation}
    \min_{\alpha} d_{\text{surf}}(\mathbf{c}_{\text{parent}} + \alpha \mathbf{d}_{\text{grow}}),
\end{equation}
where $d_{\text{surf}}$ computes the distance to the primitive's surface. 
This optimization problem is coded and exposed as a high-level API call to the VLM.

\begin{figure}
    \centering
    \setlength{\tabcolsep}{1pt}
    \renewcommand{\arraystretch}{0.7}
    \begin{tabular}{c c c c}
    \begin{overpic}[width=0.120\textwidth]{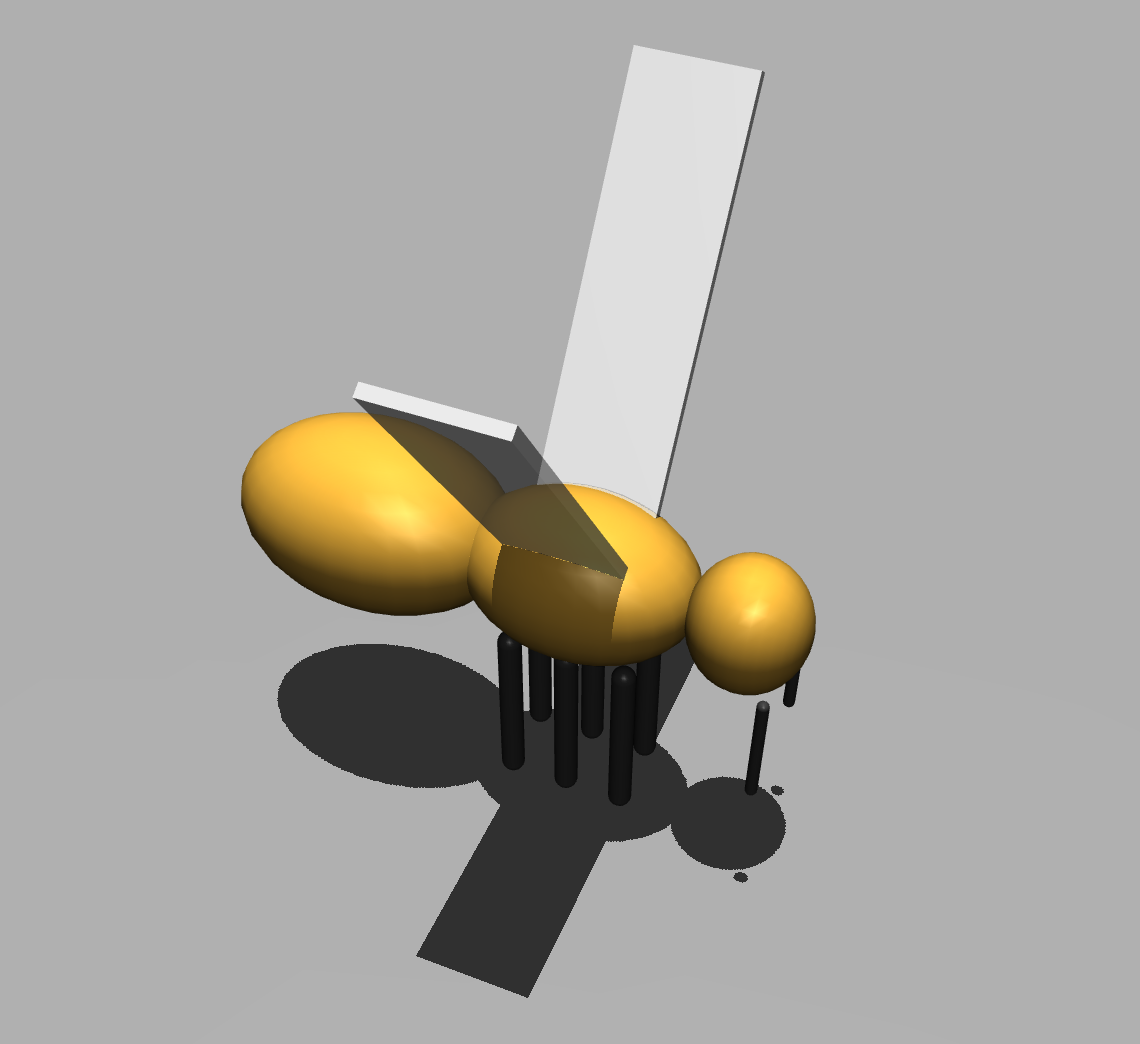}
      \put(50,80){\makebox(0,0){\scriptsize{\textcolor{black}{Bee}}}}
    \end{overpic} &
    \begin{overpic}[width=0.120\textwidth]{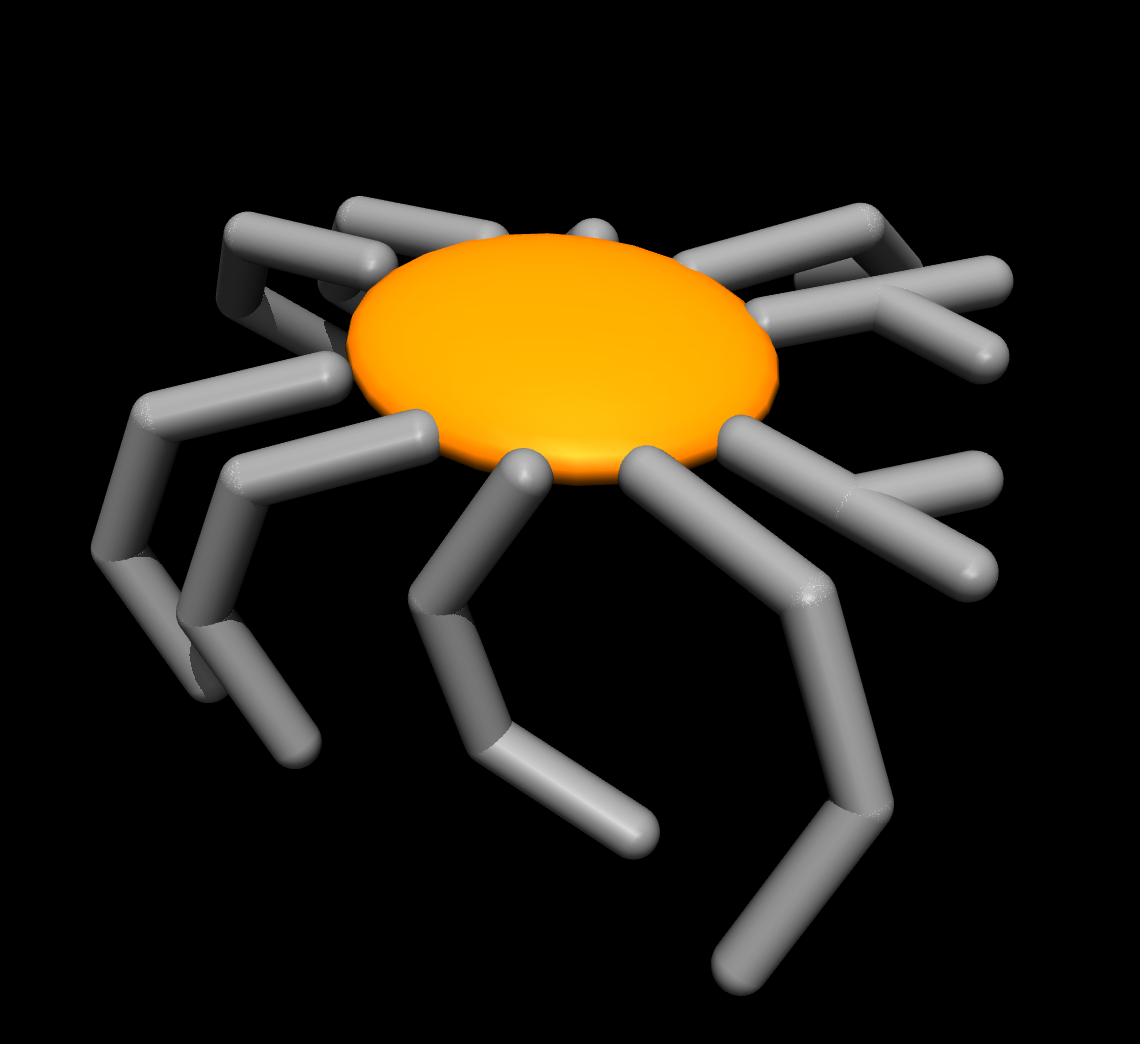}
      \put(50,80){\makebox(0,0){\scriptsize{\textcolor{white}{Crab}}}}
    \end{overpic} &
    \begin{overpic}[width=0.120\textwidth]{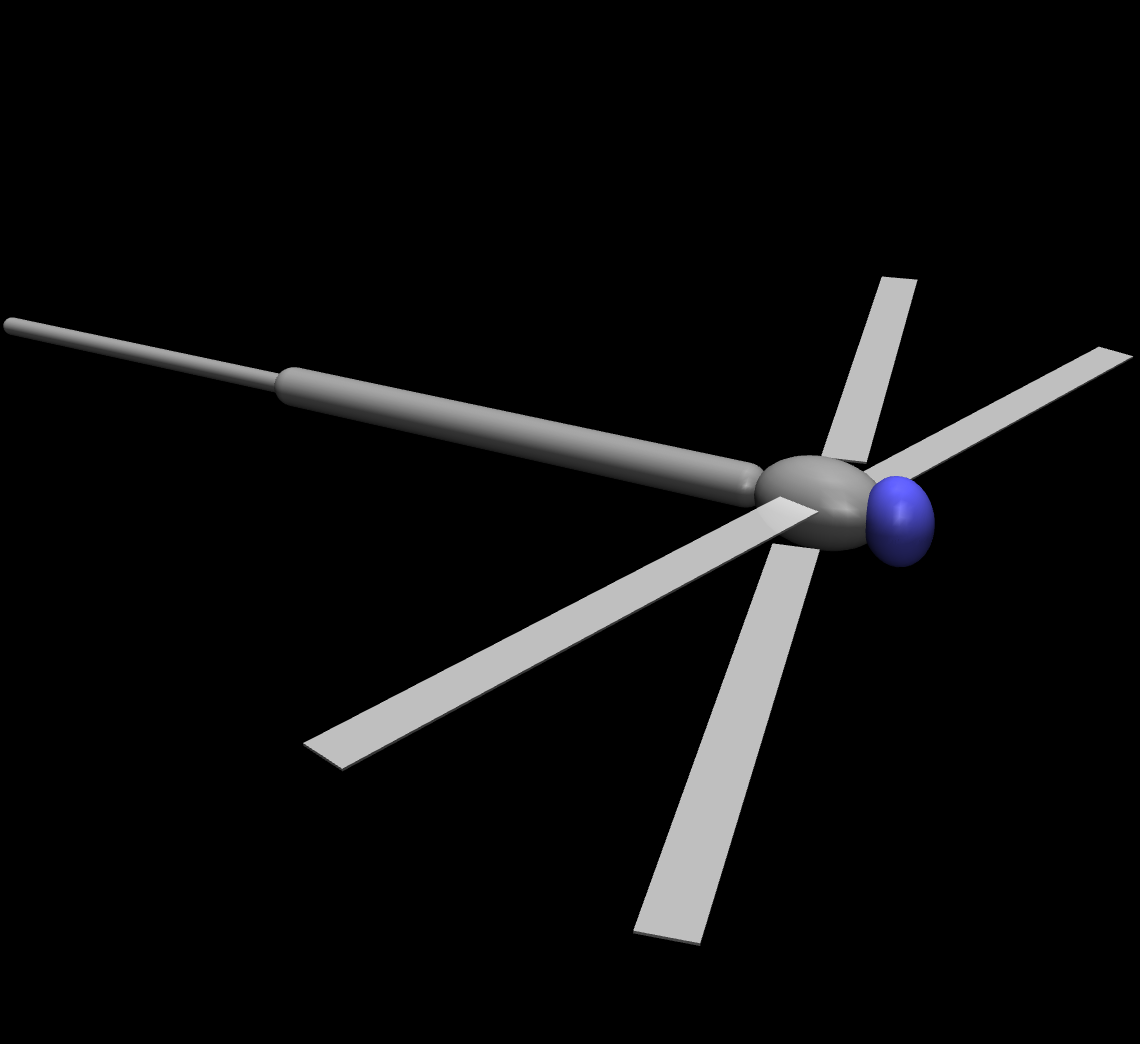}
      \put(50,80){\makebox(0,0){\scriptsize{\textcolor{white}{Dragonfly}}}}
    \end{overpic} &
    \begin{overpic}[width=0.120\textwidth]{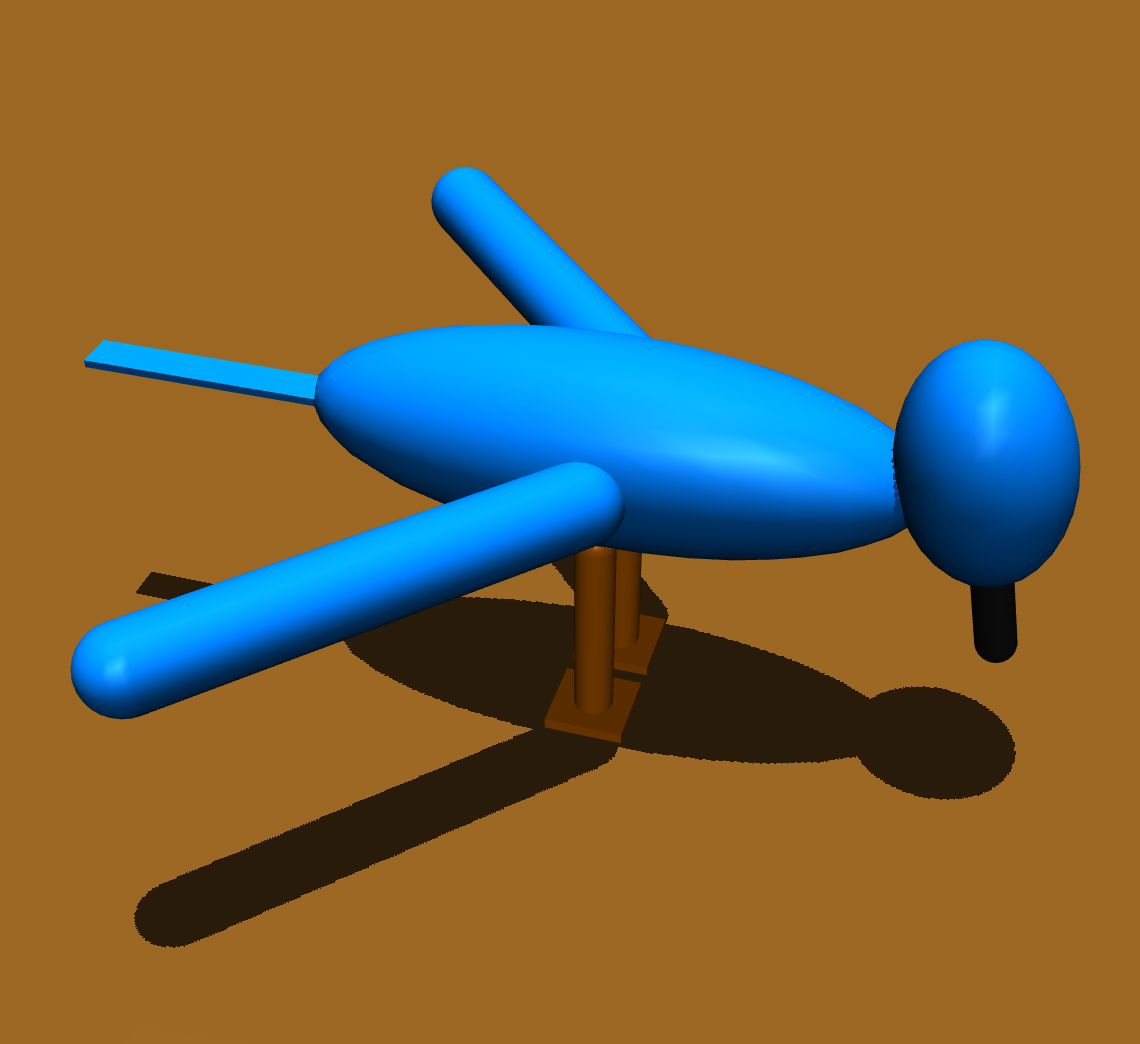}
      \put(50,80){\makebox(0,0){\scriptsize{\textcolor{black}{Kingfisher}}}}
    \end{overpic} \\
    \begin{overpic}[width=0.120\textwidth]{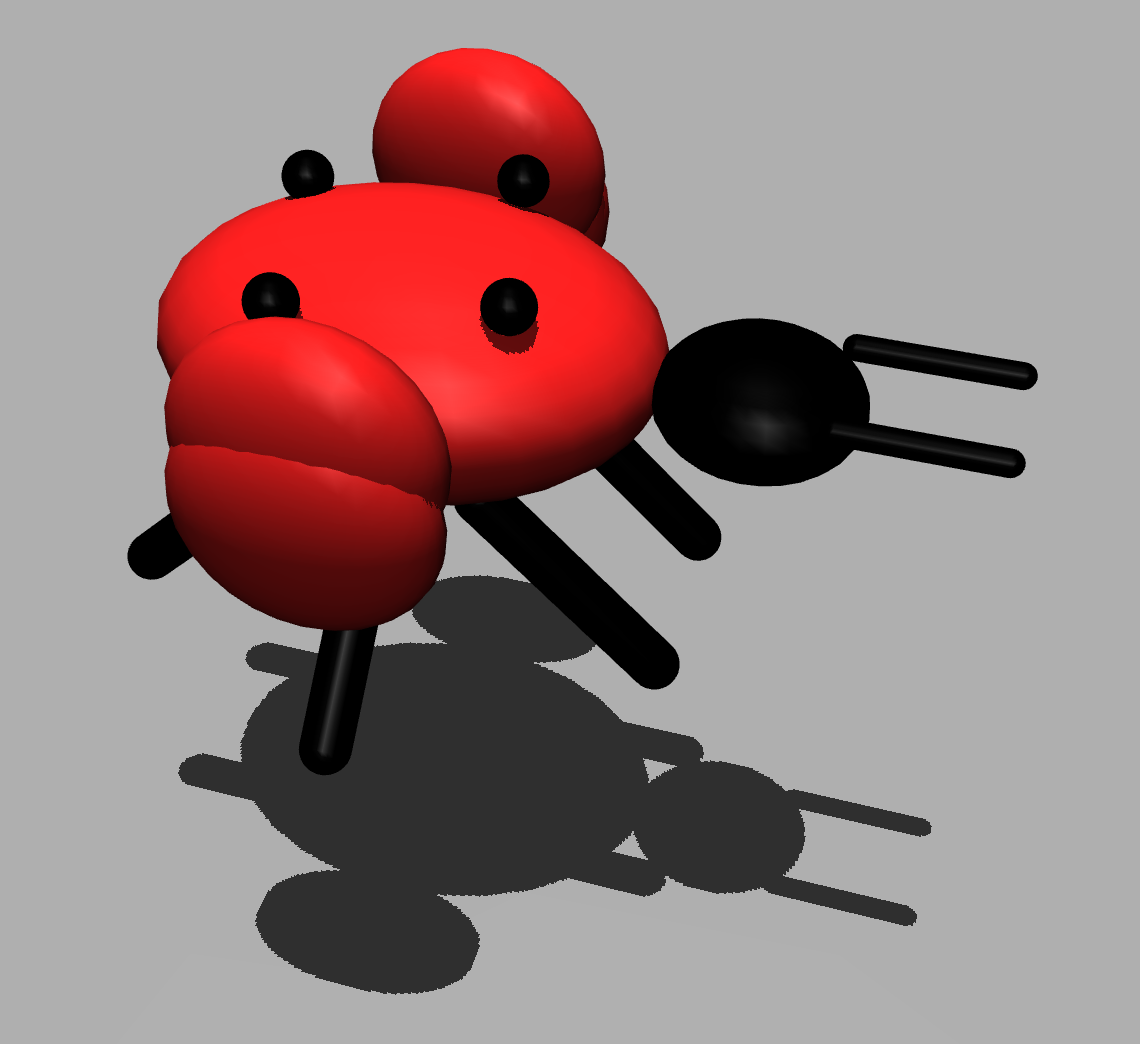}
      \put(50,80){\makebox(0,0){\scriptsize{\textcolor{black}{Ladybug}}}}
    \end{overpic} &
    \begin{overpic}[width=0.120\textwidth]{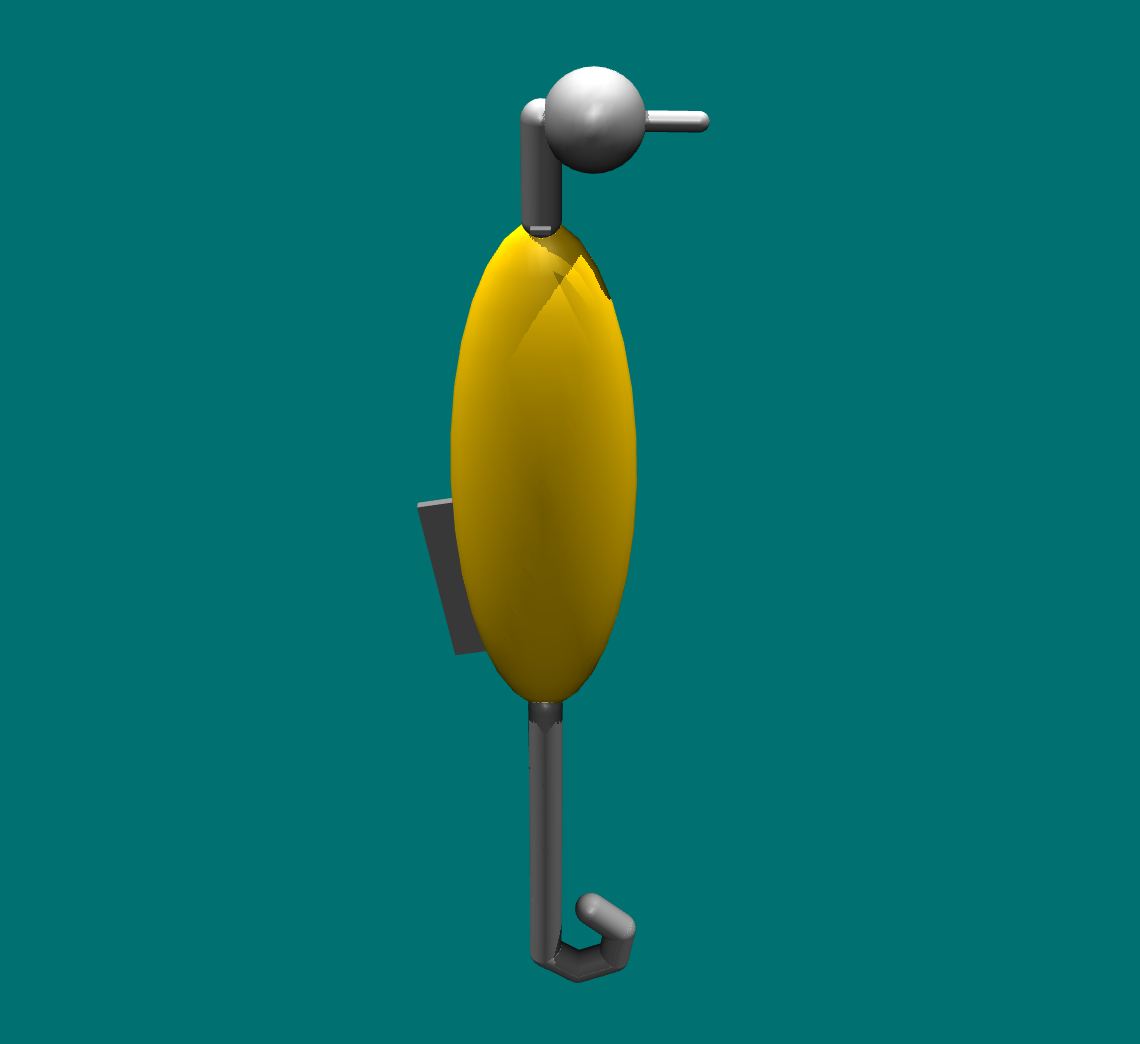}
      \put(50,80){\makebox(0,0){\scriptsize{\textcolor{black}{Seahorse}}}}
    \end{overpic} &
    \begin{overpic}[width=0.120\textwidth]{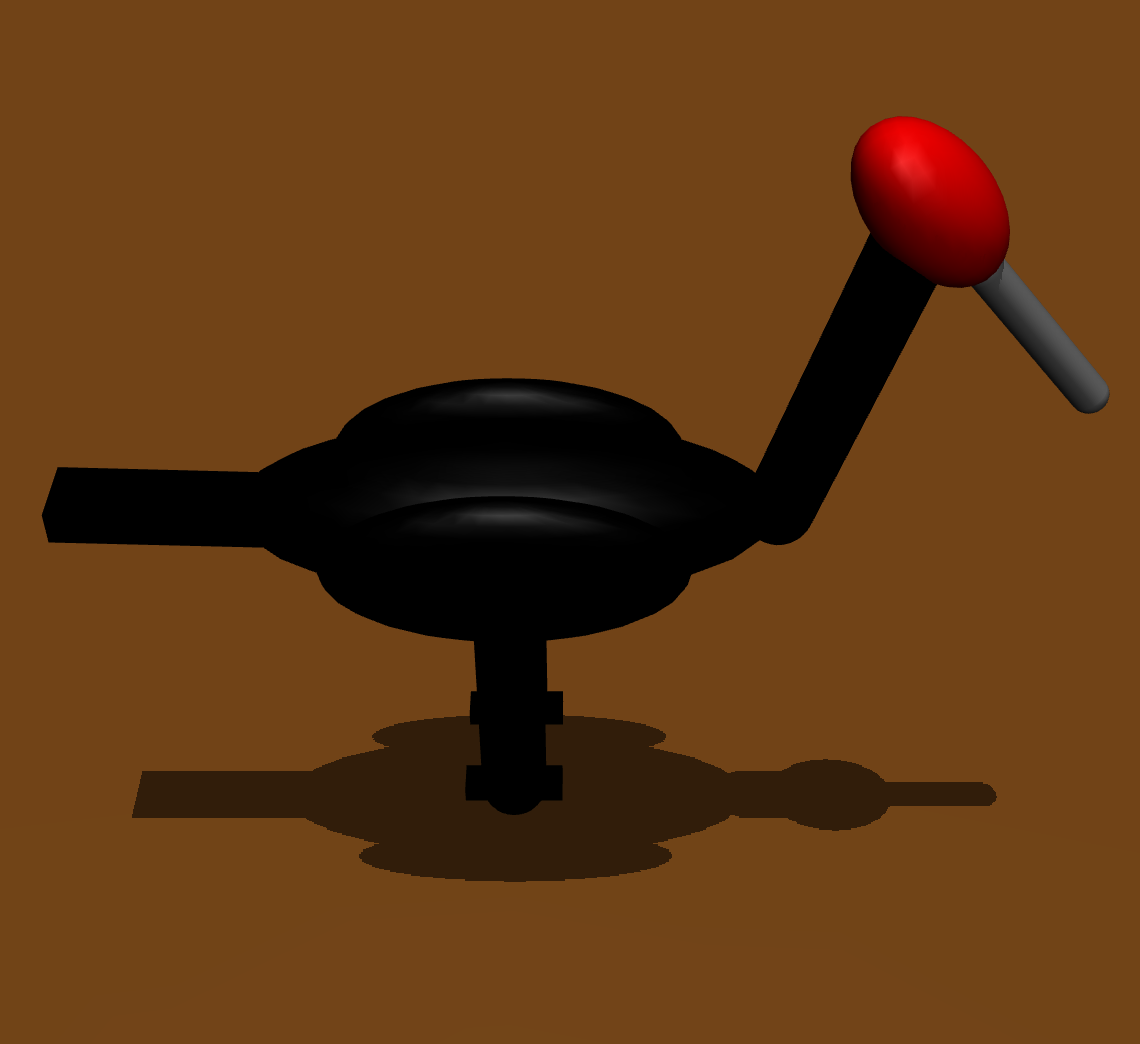}
      \put(50,80){\makebox(0,0){\scriptsize{\textcolor{black}{Woodpecker}}}}
    \end{overpic} &
    \begin{overpic}[width=0.120\textwidth]{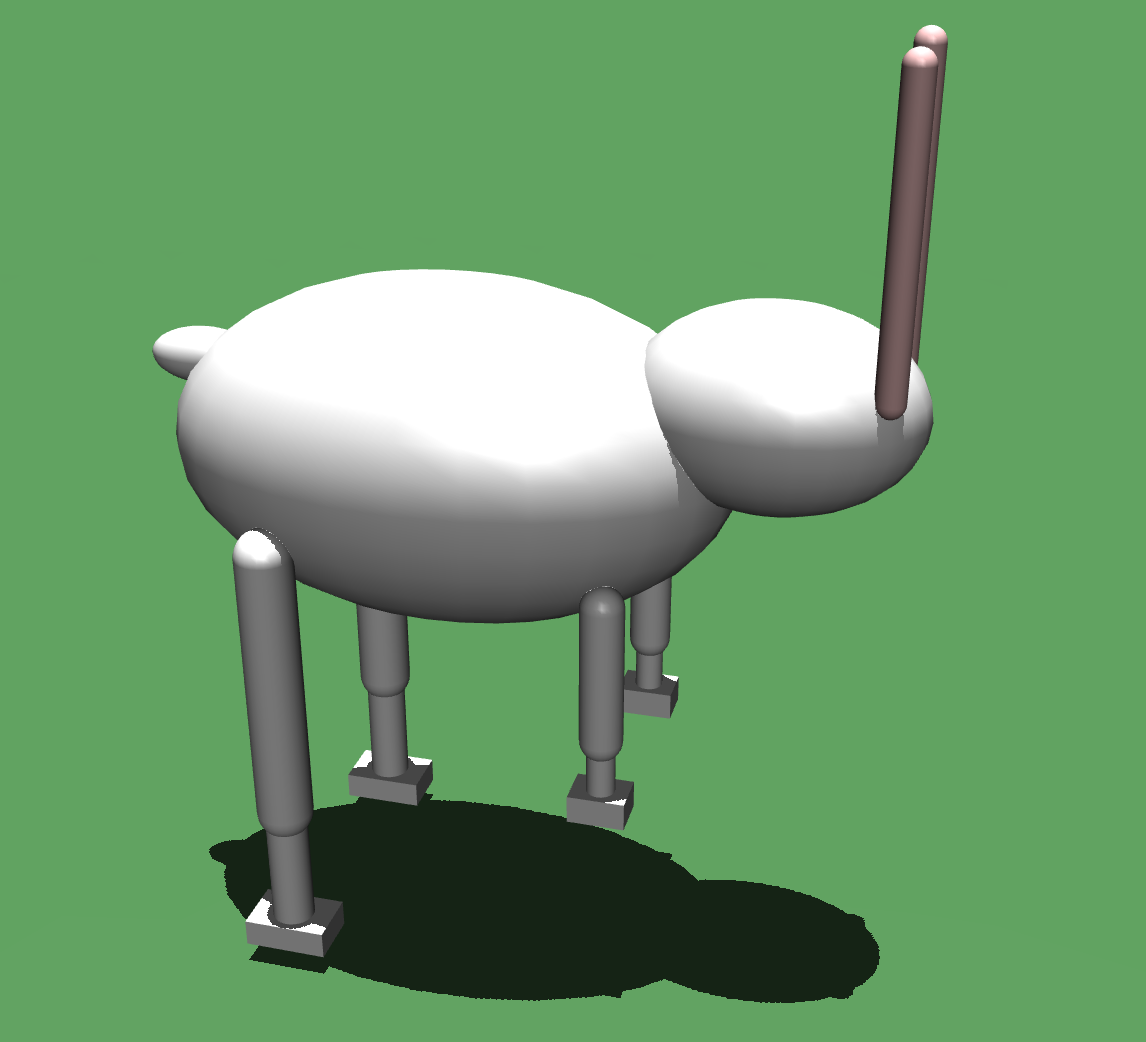}
      \put(50,80){\makebox(0,0){\scriptsize{\textcolor{black}{Rabbit}}}}
    \end{overpic} \\
    \begin{overpic}[width=0.120\textwidth]{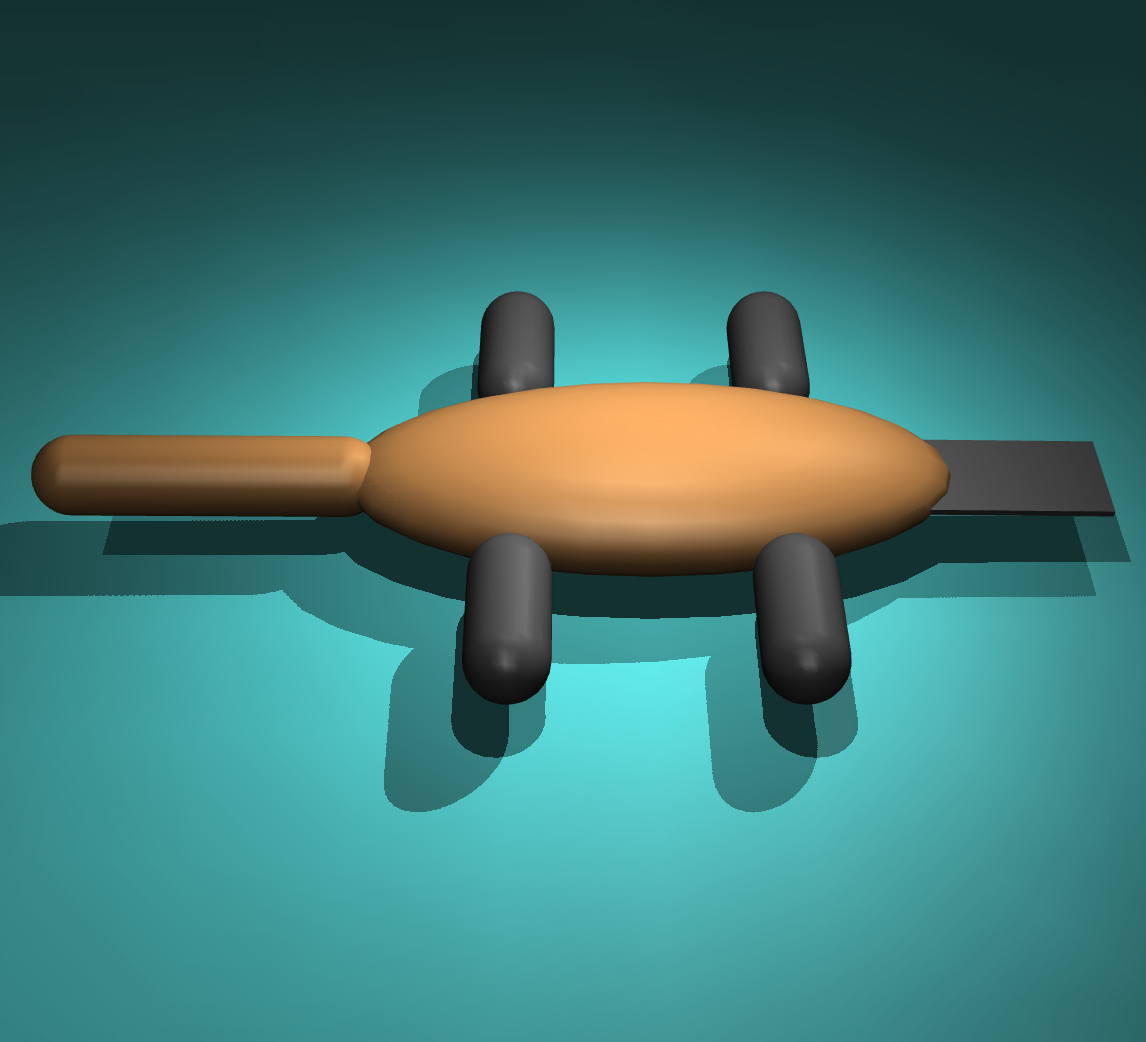}
      \put(50,80){\makebox(0,0){\scriptsize{\textcolor{white}{Platypus}}}}
    \end{overpic} &
    \begin{overpic}[width=0.120\textwidth]{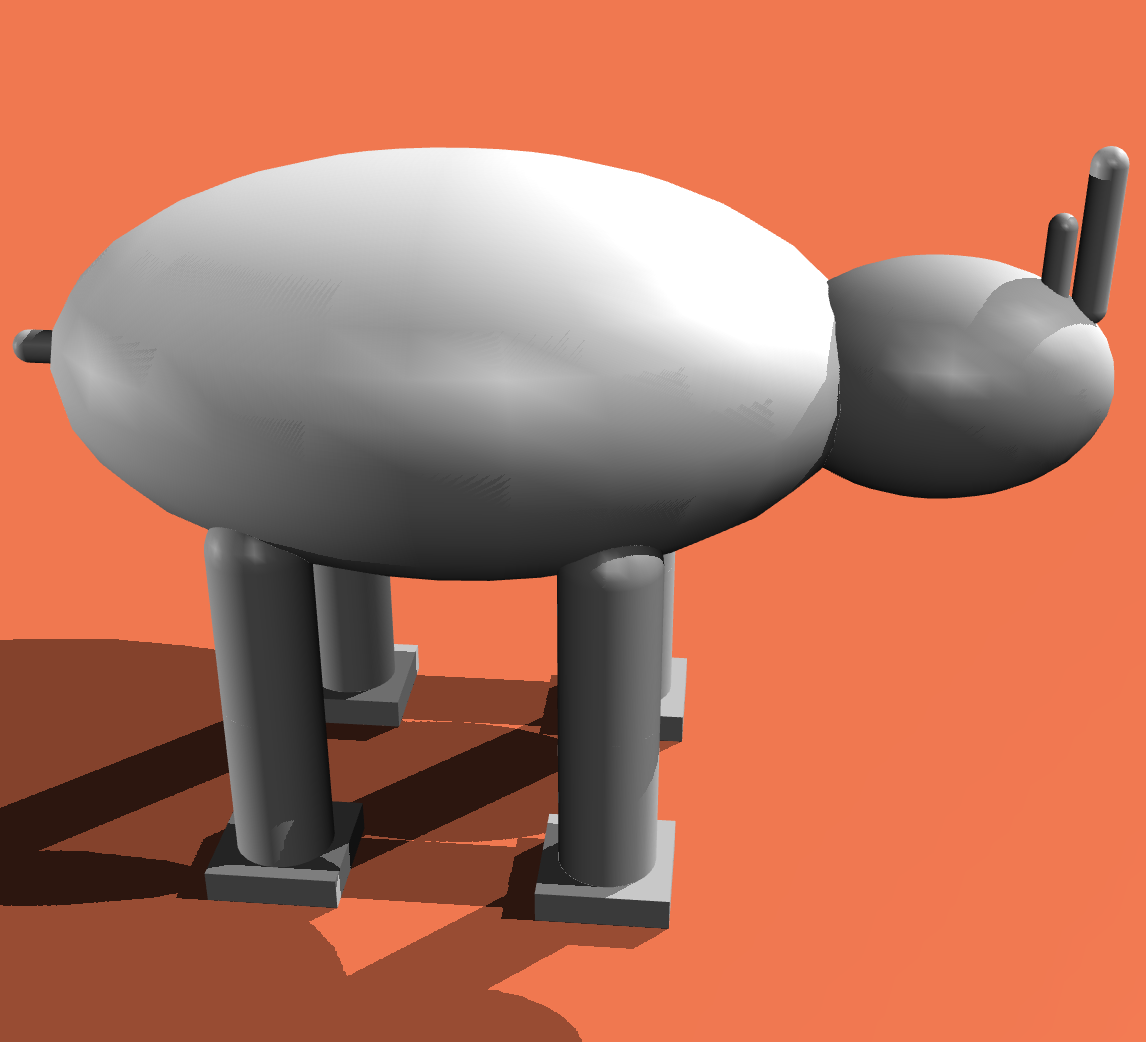}
      \put(50,80){\makebox(0,0){\scriptsize{\textcolor{black}{Rhinoceros}}}}
    \end{overpic} &
    \begin{overpic}[width=0.120\textwidth]{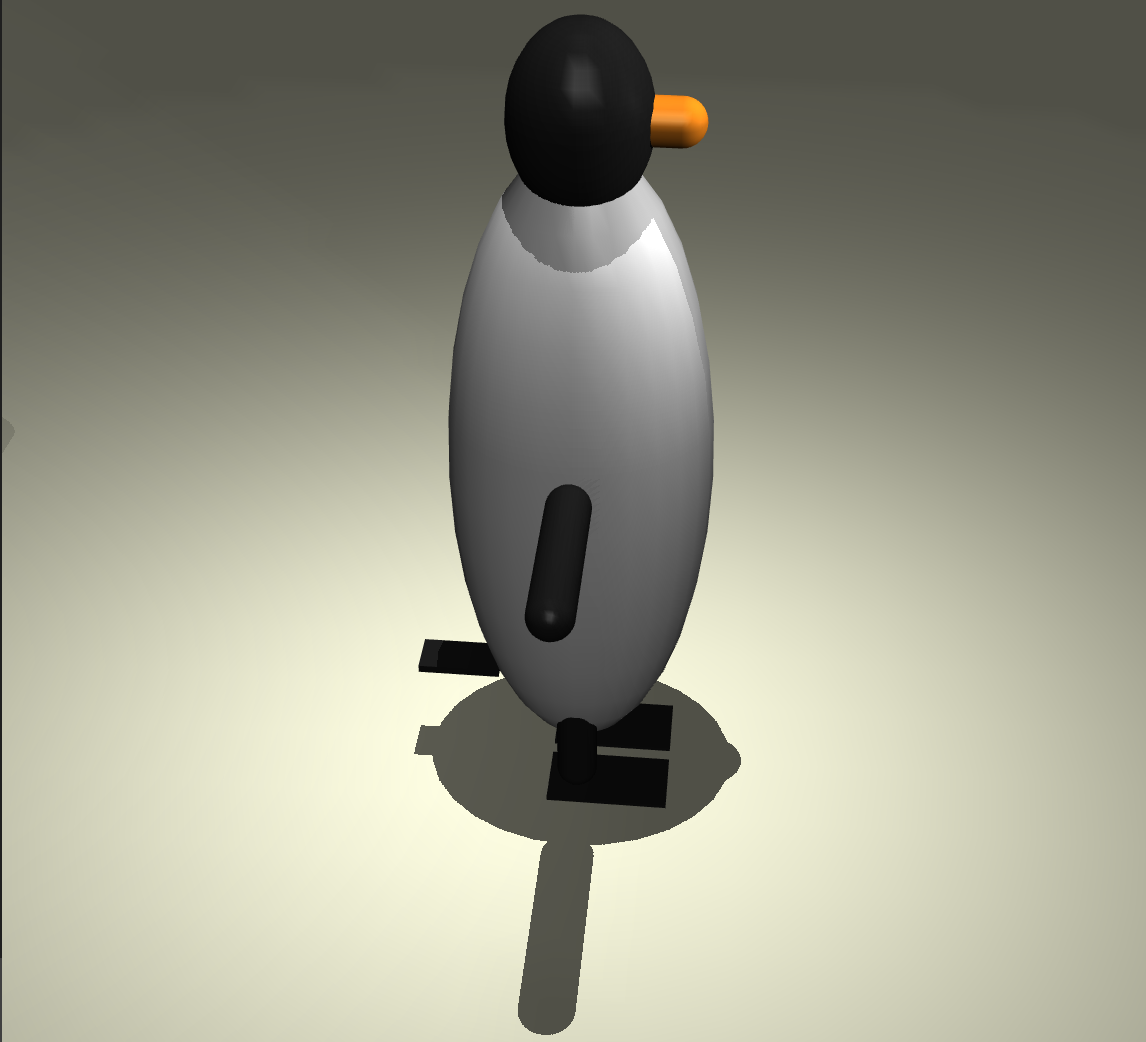}
      \put(50,80){\makebox(0,0){\scriptsize{\textcolor{white}{Penguin}}}}
    \end{overpic} &
    \begin{overpic}[width=0.120\textwidth]{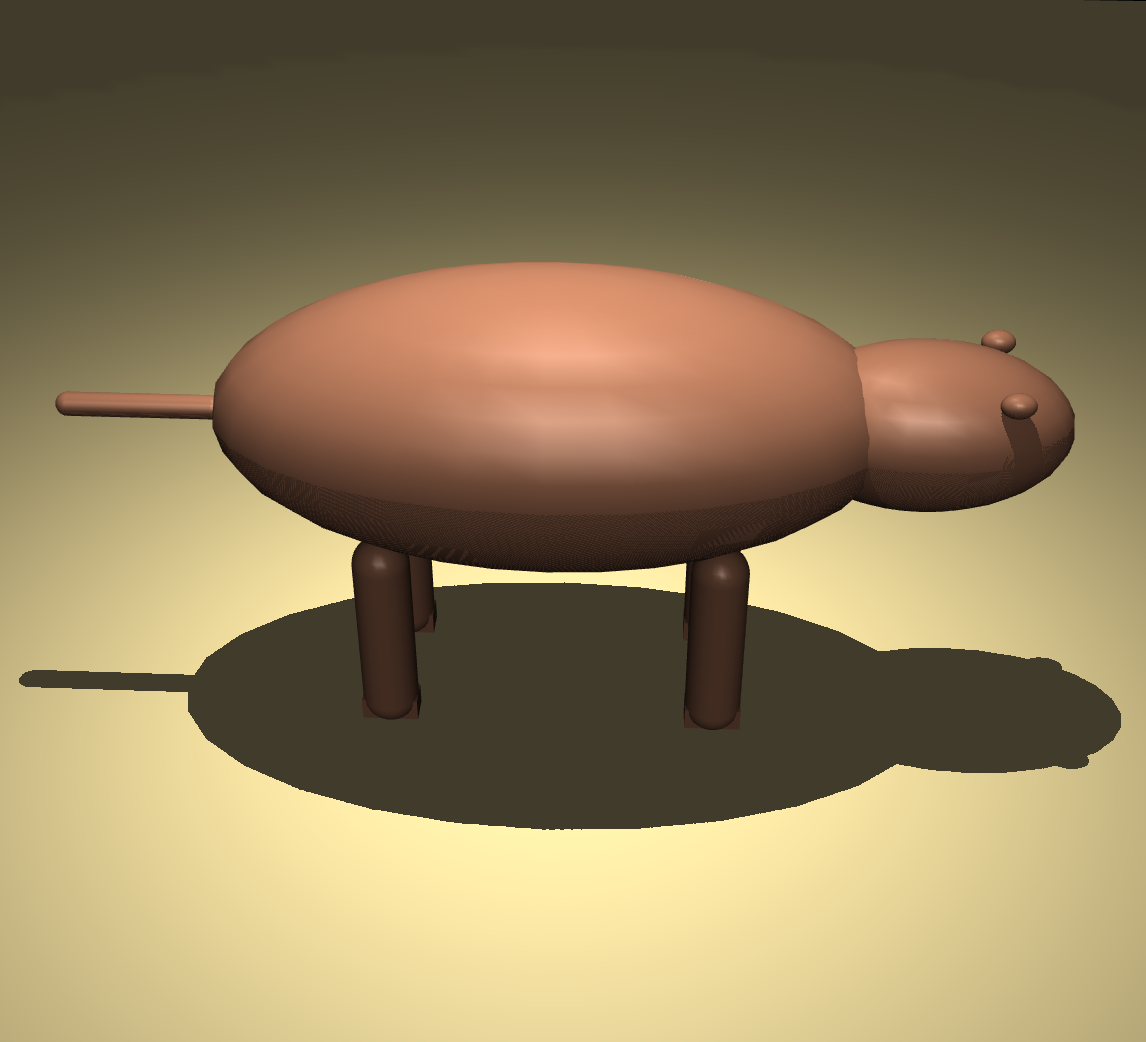}
      \put(50,80){\makebox(0,0){\scriptsize{\textcolor{black}{Hippopotamus}}}}
    \end{overpic} \\
    \begin{overpic}[width=0.120\textwidth]{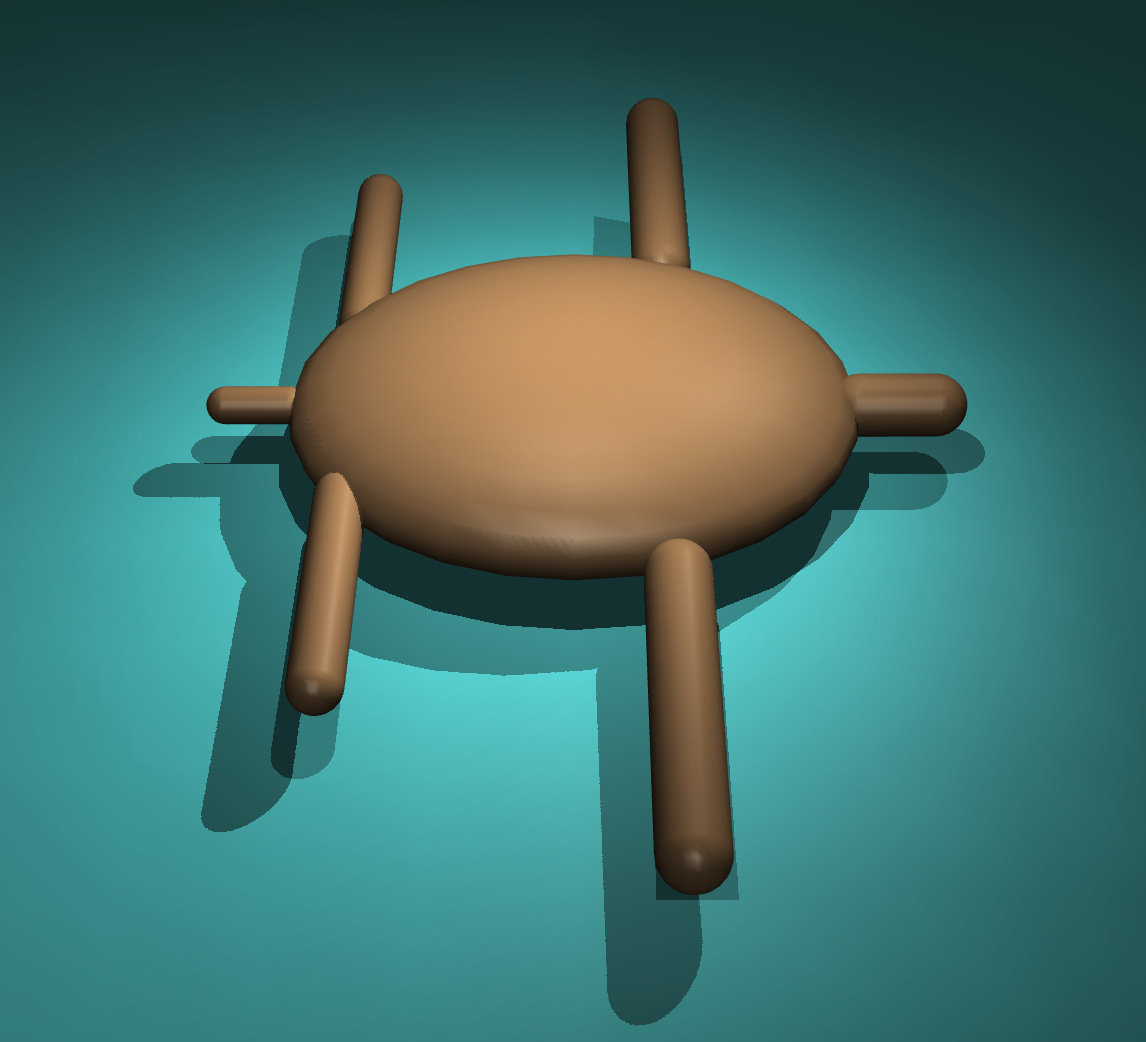}
      \put(50,80){\makebox(0,0){\scriptsize{\textcolor{black}{Turtle}}}}
    \end{overpic} &
    \begin{overpic}[width=0.120\textwidth]{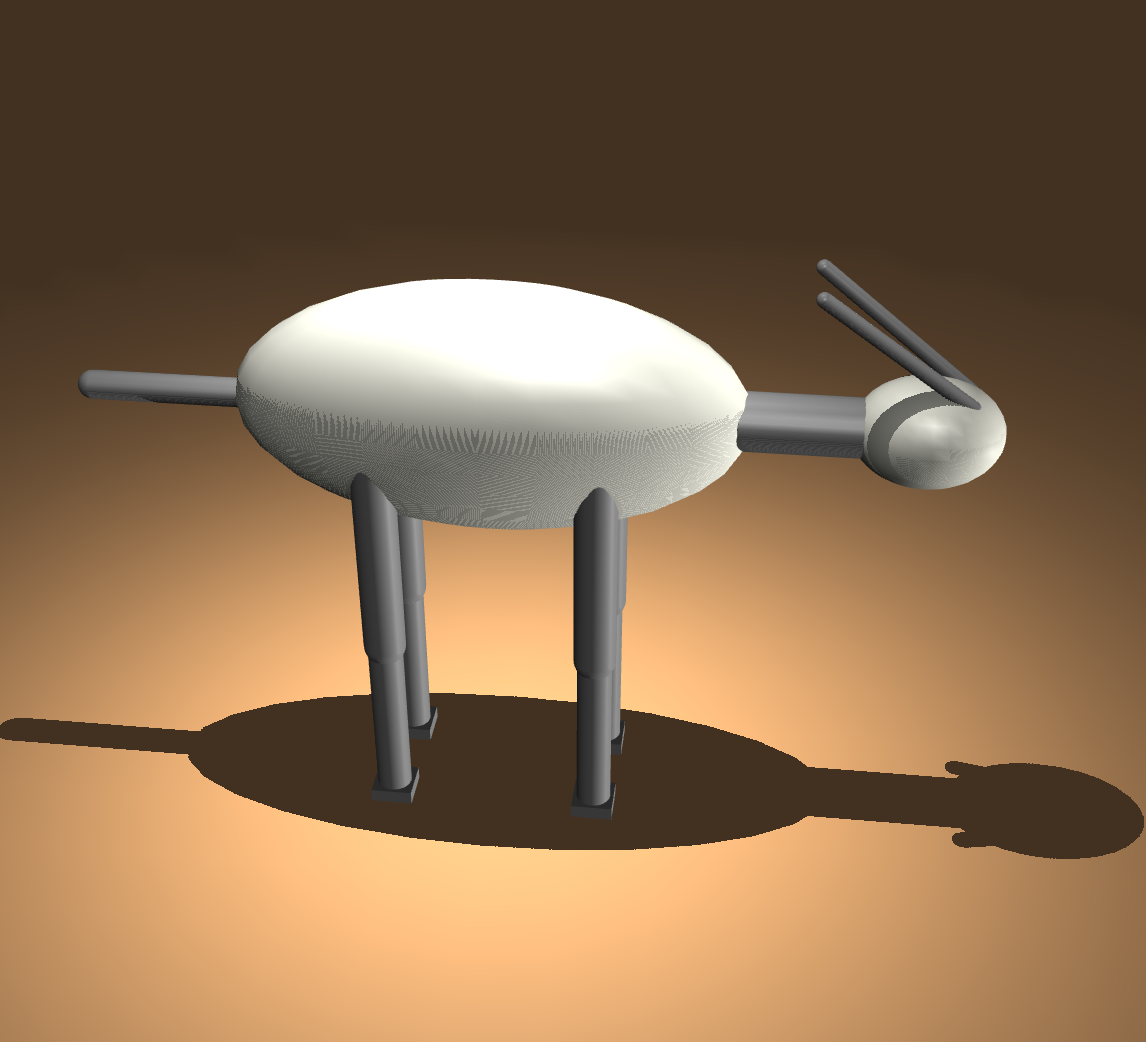}
      \put(50,80){\makebox(0,0){\scriptsize{\textcolor{white}{Addax}}}}
    \end{overpic} &
    \begin{overpic}[width=0.120\textwidth]{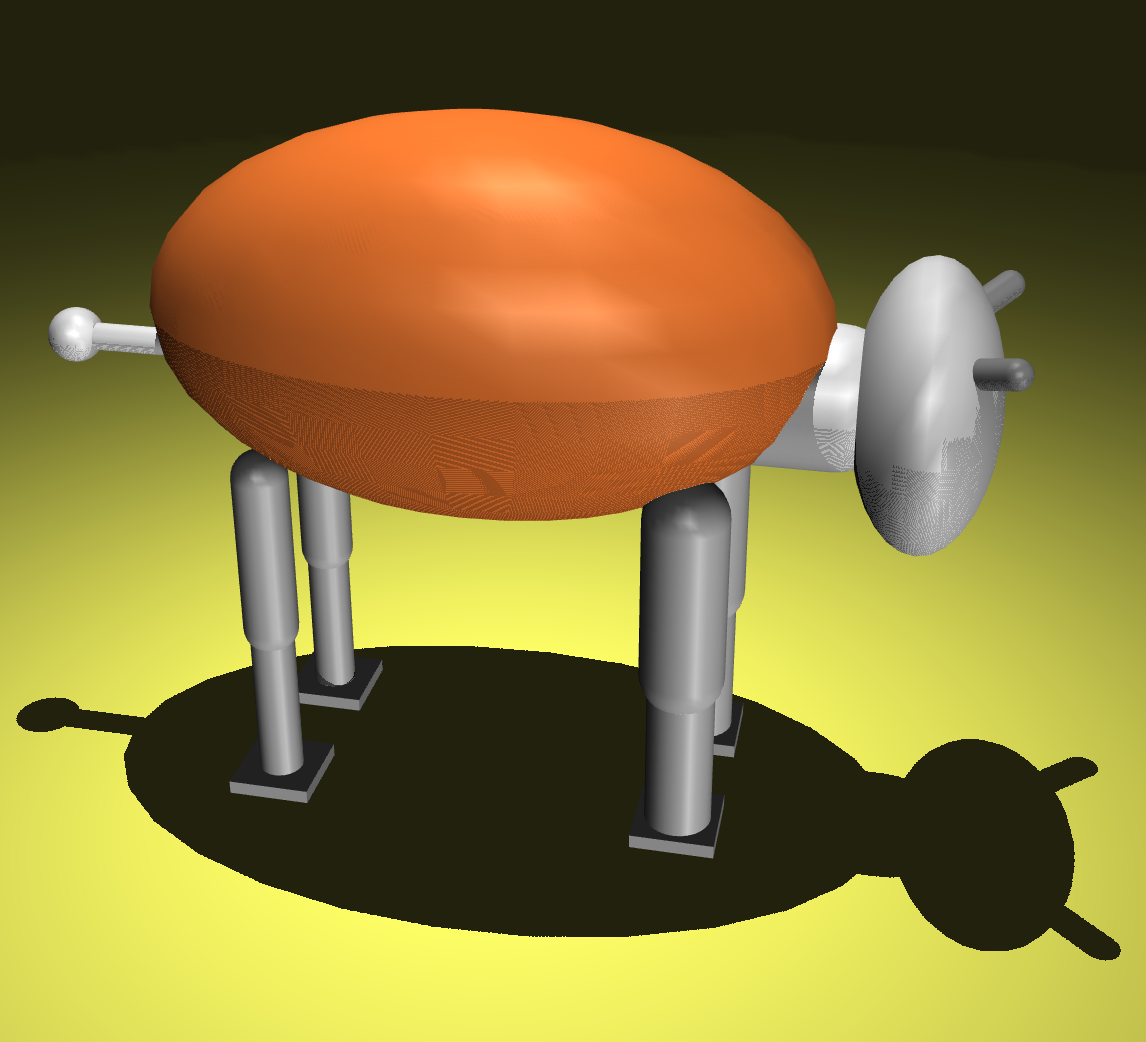}
      \put(50,80){\makebox(0,0){\scriptsize{\textcolor{gray}{Bison}}}}
    \end{overpic} &
    \begin{overpic}[width=0.120\textwidth]{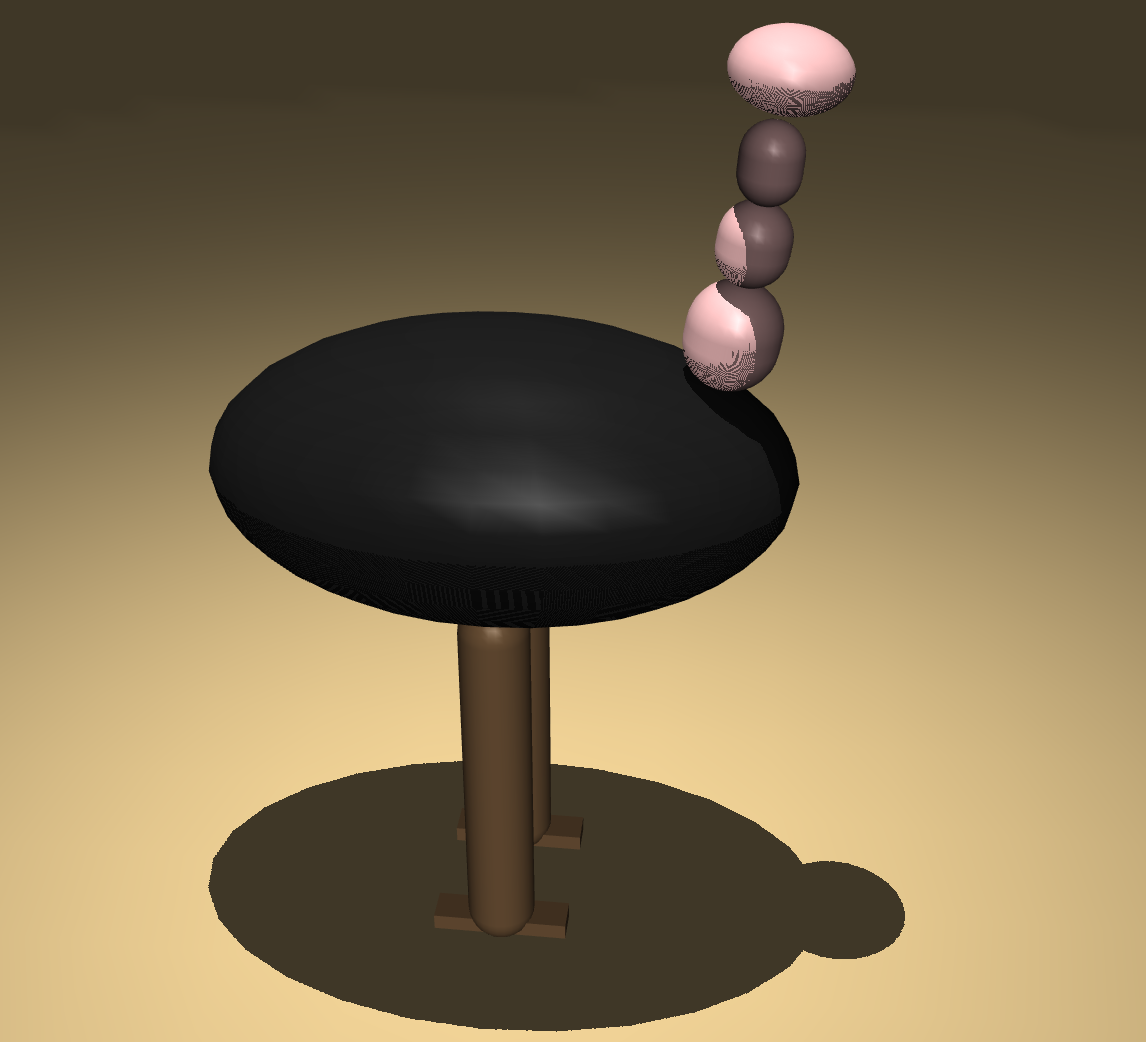}
      \put(50,80){\makebox(0,0){\scriptsize{\textcolor{white}{Ostrich}}}}
    \end{overpic} \\
    \begin{overpic}[width=0.120\textwidth]{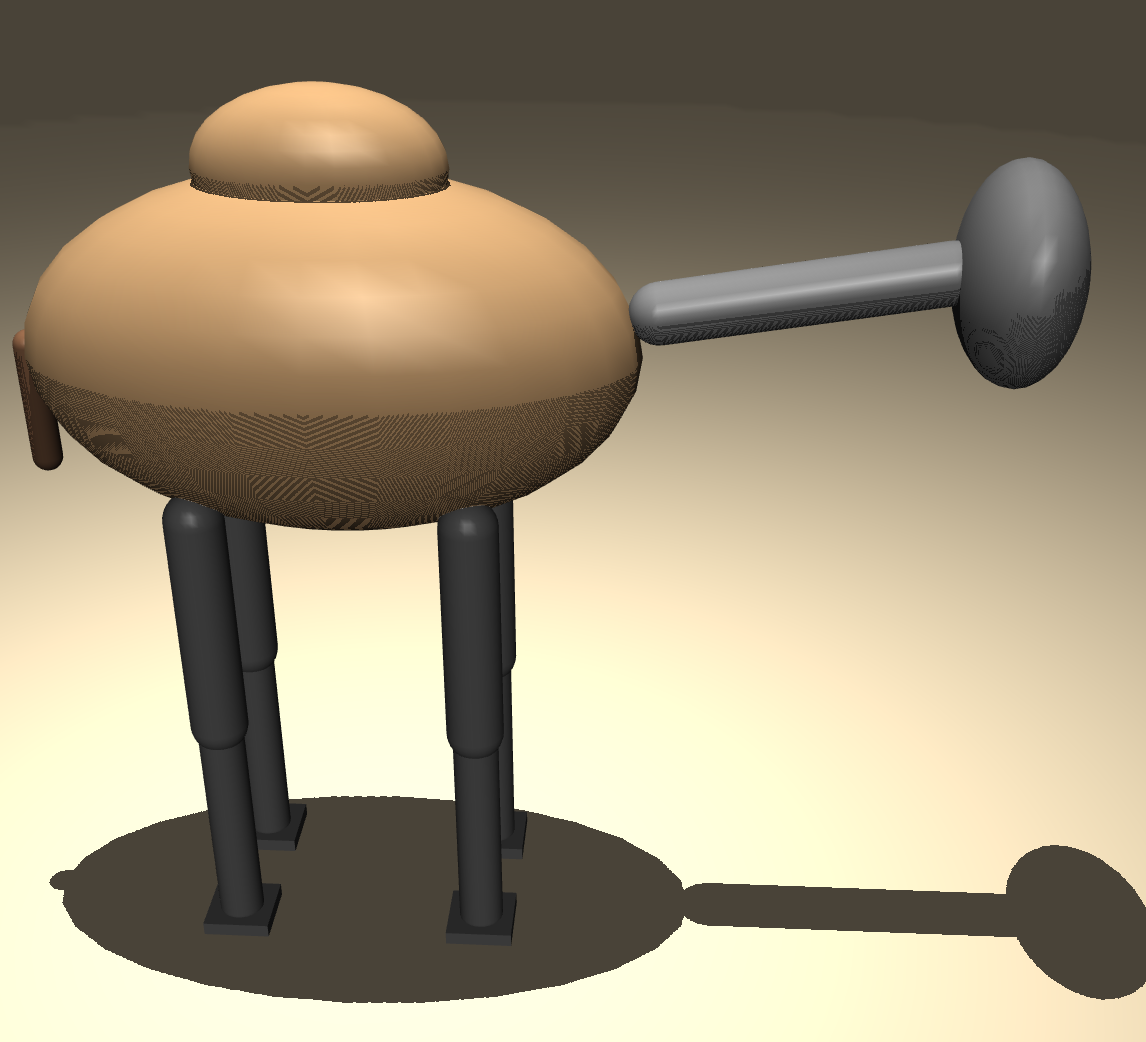}
      \put(50,80){\makebox(0,0){\scriptsize{\textcolor{gray}{Camel}}}}
    \end{overpic} &
    \begin{overpic}[width=0.120\textwidth]{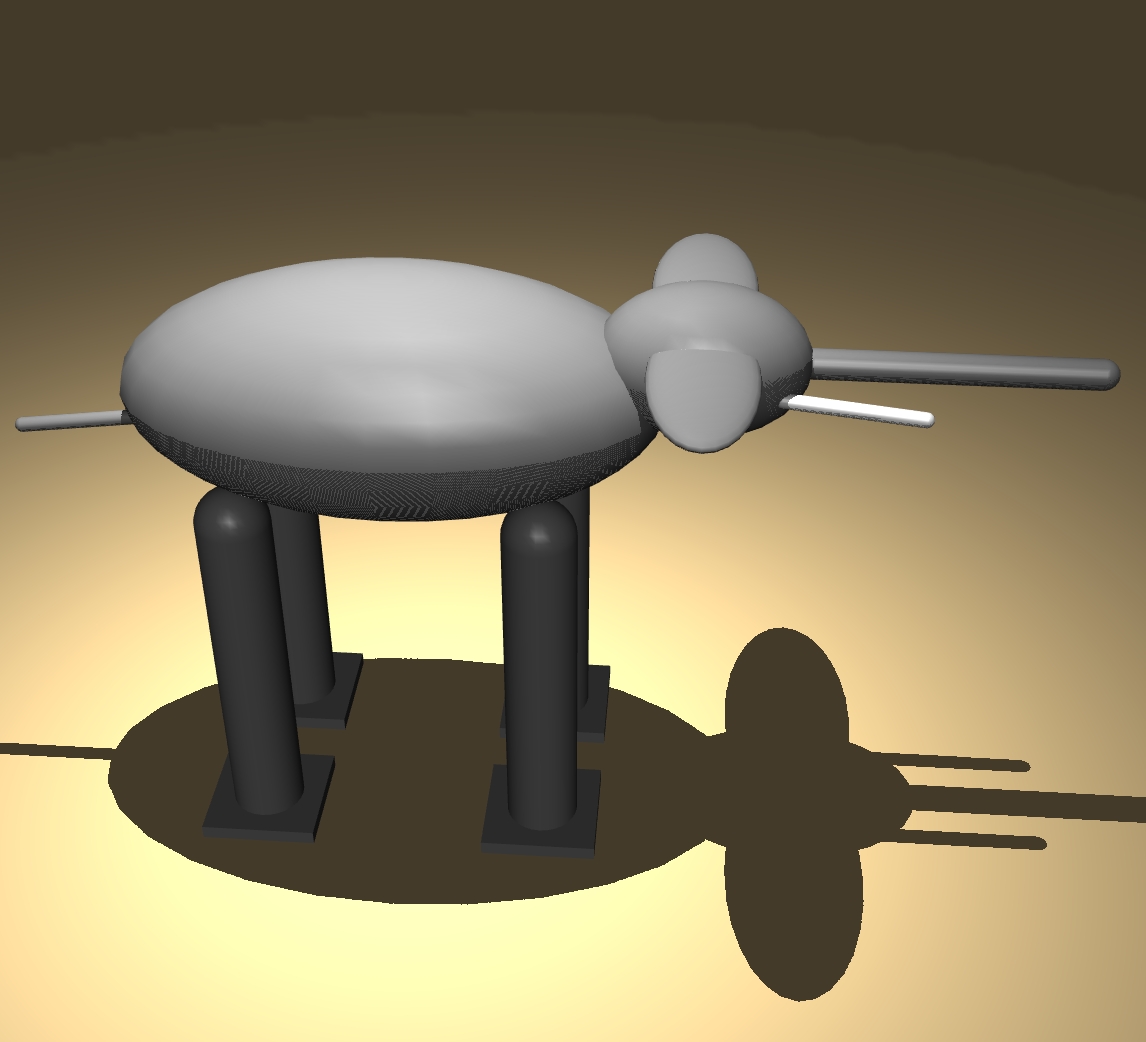}
      \put(50,80){\makebox(0,0){\scriptsize{\textcolor{gray}{Elephant}}}}
    \end{overpic} &
    \begin{overpic}[width=0.120\textwidth]{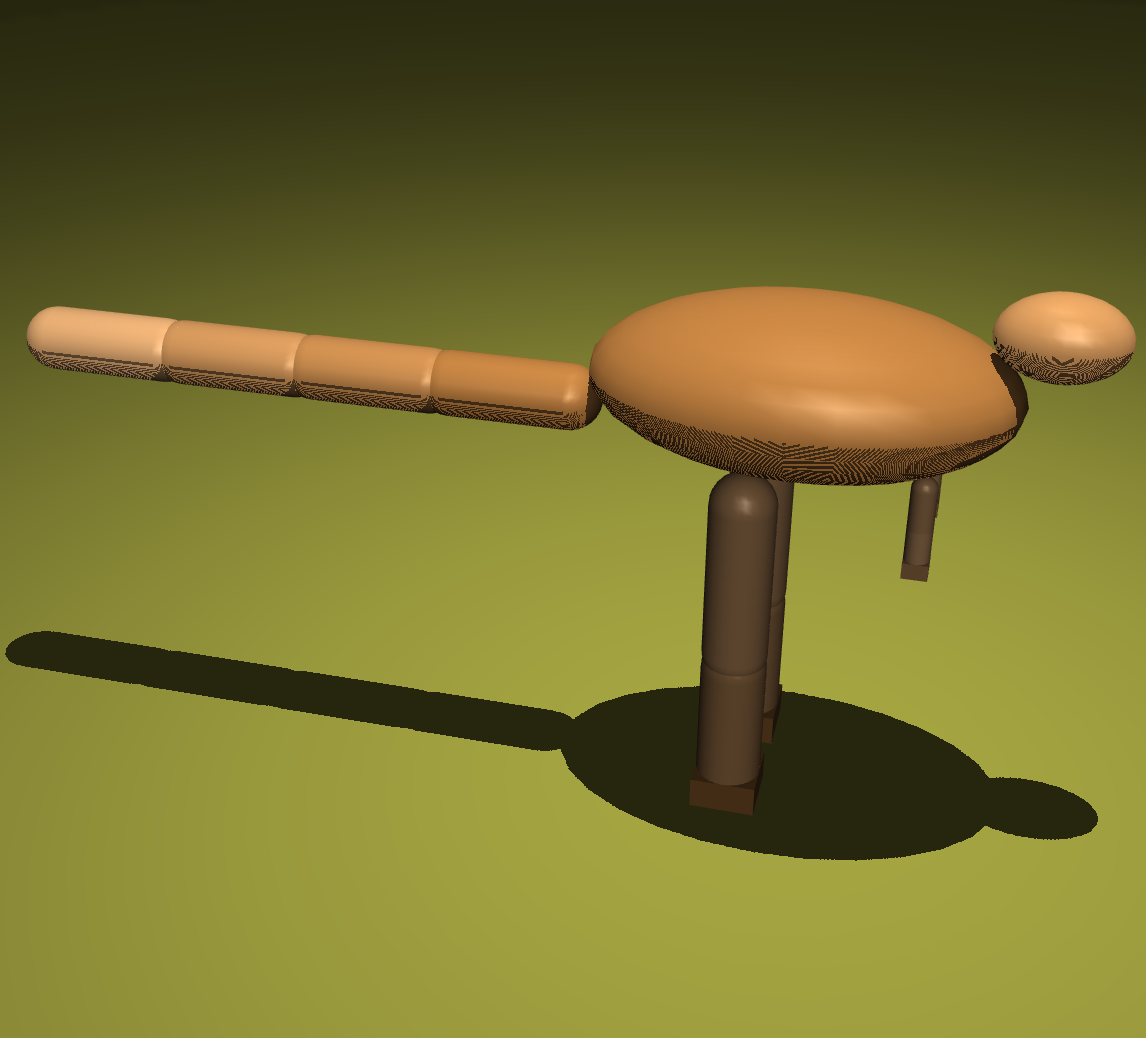}
      \put(50,80){\makebox(0,0){\scriptsize{\textcolor{white}{Kangaroo}}}}
    \end{overpic} &
    \begin{overpic}[width=0.120\textwidth]{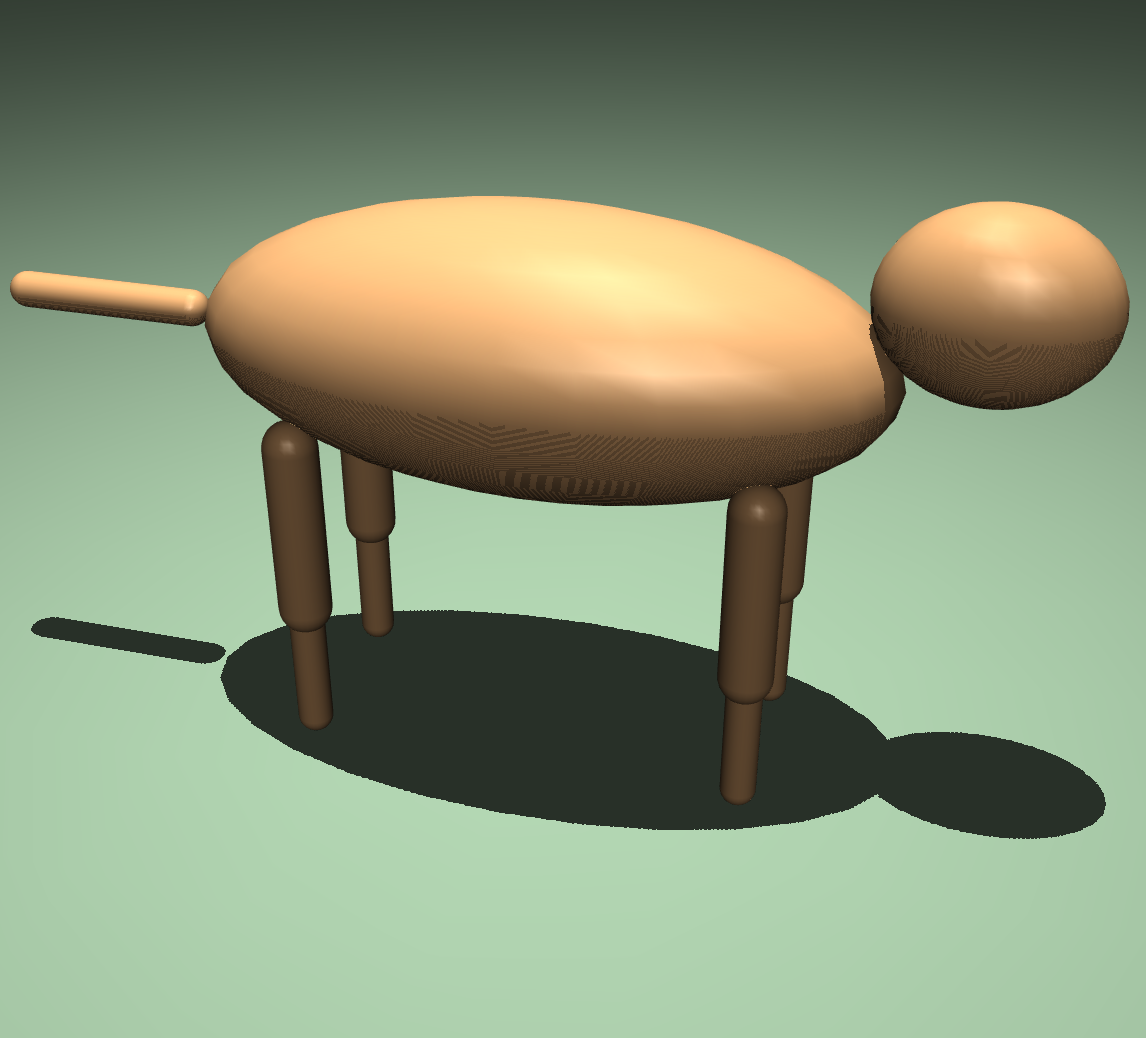}
      \put(50,80){\makebox(0,0){\scriptsize{\textcolor{black}{Dog}}}}
    \end{overpic} \\
    \end{tabular}

    \caption{Articulated robot models created by the proposed method, which include all of the land, sea, and air creatures.}
    \label{fig:model_creation}	
\end{figure}

\noindent \textbf{Joint Type.}
The framework then determines the type of joint, selecting from hinge, ball, or free. If it is a hinge joint, the framework must also determine the axis. To provide the VLM with proper context, we give more detailed instructions, including information about the front direction or an up vector.

\noindent \textbf{Geometric Primitives.}
Next, the framework determines an associated geometric primitive. We limit each node to a single geometric primitive. The VLM is instructed to choose the shape of the primitive from options like box, ellipsoid, and capsule, and to determine the geometric properties of the new component.


\noindent \textbf{Body Dimensions.}
Finally, the framework determines the body dimensions. Since we assume a fully connected robot without any penetration or gap, the body dimensions are automatically computed based on the size of the geometric primitive using its bounding box.
\subsection{\textbf{Automated Visual Feedback}}
By completing steps 1 and 2, the framework can synthesize a valid robot description file. However, the resulting design may deviate from the user's expected visual appearance. For example, while the generated design may include all the structural elements of an elephant, the proportions between the body and legs might not be properly adjusted - please refer to the result section for illustrative scenarios. To address this, we implement a third step: fine-tuning the generated design based on a user-provided reference image. An image of the model is rendered and used as part of the input context along with the model definition. The VLM is then instructed to compare the rendered view with the reference image and modify the model to better match the reference. This process is repeated multiple times until the user is satisfied with the results. In our experiments, we obtain visually acceptable results in three iterations.




\subsection{\textbf{Human Feedback}}
The final step is human feedback, which is similar to the previous visual feedback step, but it incorporates user language feedback as input. This step allows users to easily edit the generated designs using natural language instructions to match their preferences. The number of prompts is also limited to three. We note that in many cases, the model's fidelity after the visual refinement stage is sufficiently high, and user modifications are unnecessary. However, in some instances, user feedback can further enhance visual fidelity. Examples of such prompts include `Make the legs shorter,' `Make the wings flatter,' and `Make the horns point backward.'

\section{EXPERIMENTS AND RESULTS}

We aim to answer the following research questions:
\begin{itemize}
    \item Can our proposed \textbf{RobotDesignGPT} generate valid and physically functional robot models?
    \item How does visual feedback affect the design quality?
    \item How does human feedback affect the design quality?
\end{itemize}

\subsection{Experimental Setup}
In all of our experiments, we employed GPT-4o~\cite{hurst2024gpt} as our VLM of choice. We set the `temperature' parameter to $0$, `top\_p' value of 1, the `frequency penalty' to $0$, the `presence penalty' to 0, and set the number of tokens to the maximum possible value which at the time this paper was submitted was $16383$. The robot models are described using the MJCF format, native to MuJoCo \cite{todorov2012mujoco}. During the design process, the design is incrementally updated using the PyMJCF library available as part of the dm\_control package \cite{tunyasuvunakool2020dm_control}.
In Section~\ref{sec:motion}, we animate various robot models to demonstrate the functional capabilities of the generated designs. We first generate trajectories using the trajectory optimization software TOWR~\cite{winkler18} and render them using Blender~\cite{blender}.


\subsection{Main Results: Model Creation}

We first evaluated the proposed \textbf{RobotDesignGPT} on $20$ user inputs inspired by nature, including legged animals, birds, and insects, to illustrate the capability of our method to design robots with arbitrary morphologies. The number of body nodes in the robots generated ranges from seven (six joints) for a simple turtle robot to $31$ ($30$ joints) for the complicated crab robot. With minimal human feedback, the framework was able to generate visually convincing robot designs that match the user's expectations. The results are presented in Fig.~\ref{fig:model_creation}. 

\begin{figure}
    \centering
    \includegraphics[width=0.9\linewidth]{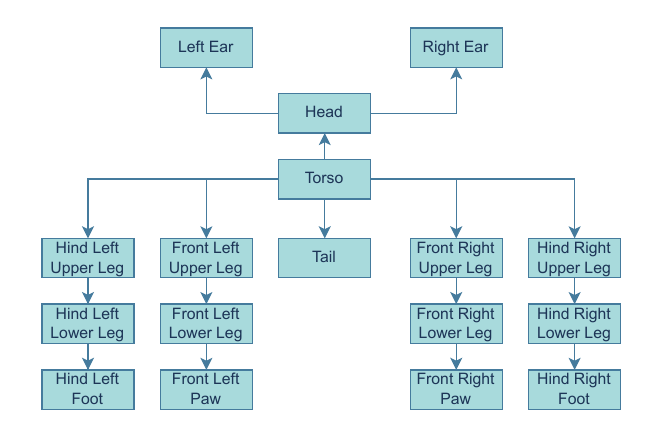}
    \caption{Kinematic tree of a rabbit robot generated by the first step of our method, structure synthesis.
    }
    \label{fig:rabbit_tree}
\end{figure}

\noindent\textbf{Kinematic tree structure.} 
We observed that the proposed multi-step design synthesis allows the framework to successfully capture the user-specified creature's structure. We illustrate one example by visualizing the kinematic tree structure for a rabbit robot in Fig.~\ref{fig:rabbit_tree}. Our framework successfully identifies key anatomical features like limbs, head, ears, and tails, properly connected in a kinematic tree.

\begin{figure}
    \centering
    \setlength{\tabcolsep}{1pt}
    \renewcommand{\arraystretch}{0.7}
    \begin{tabular}{c c c}
    \includegraphics[width=0.33\linewidth]{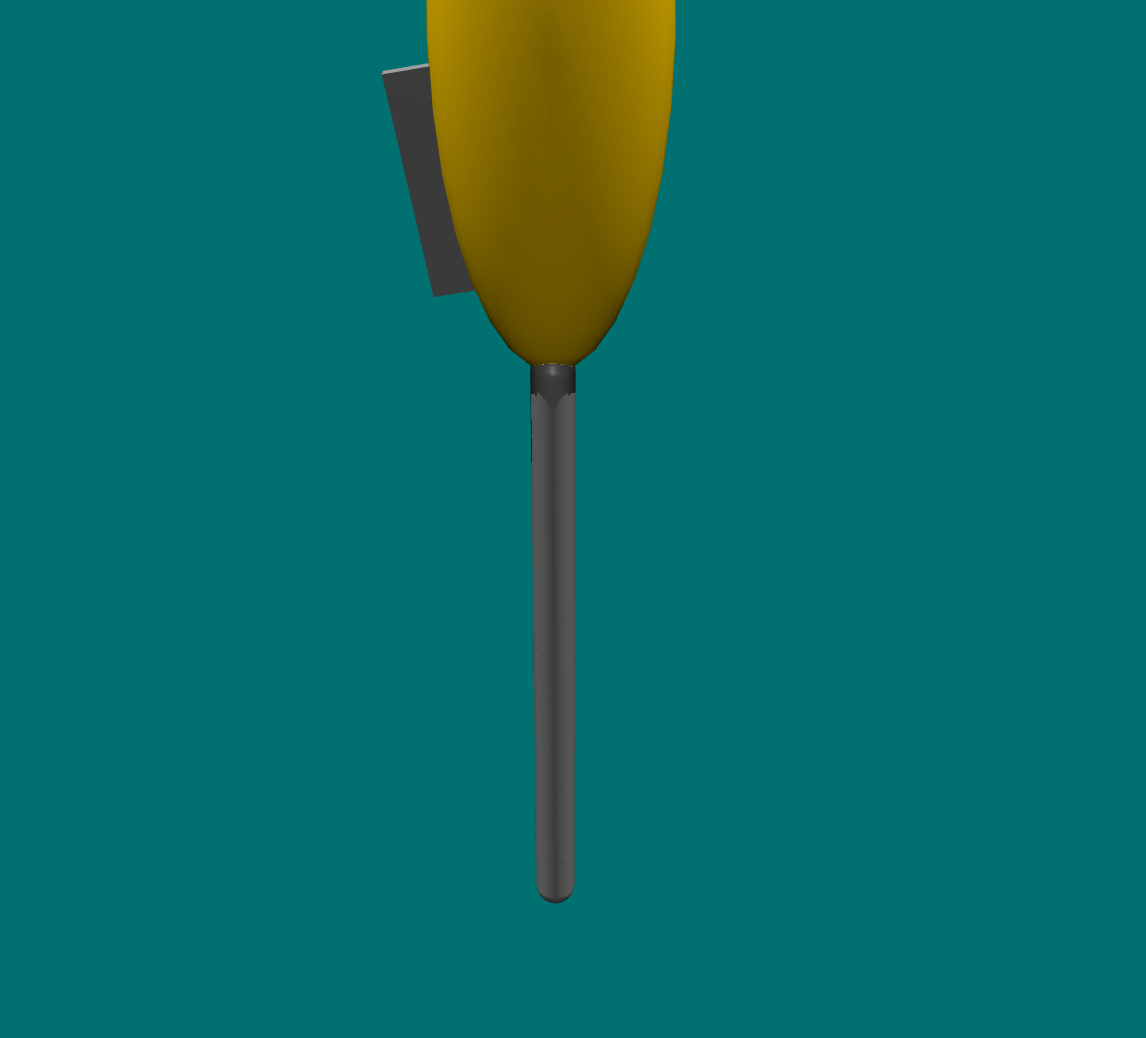} & 
    \includegraphics[width=0.33\linewidth]{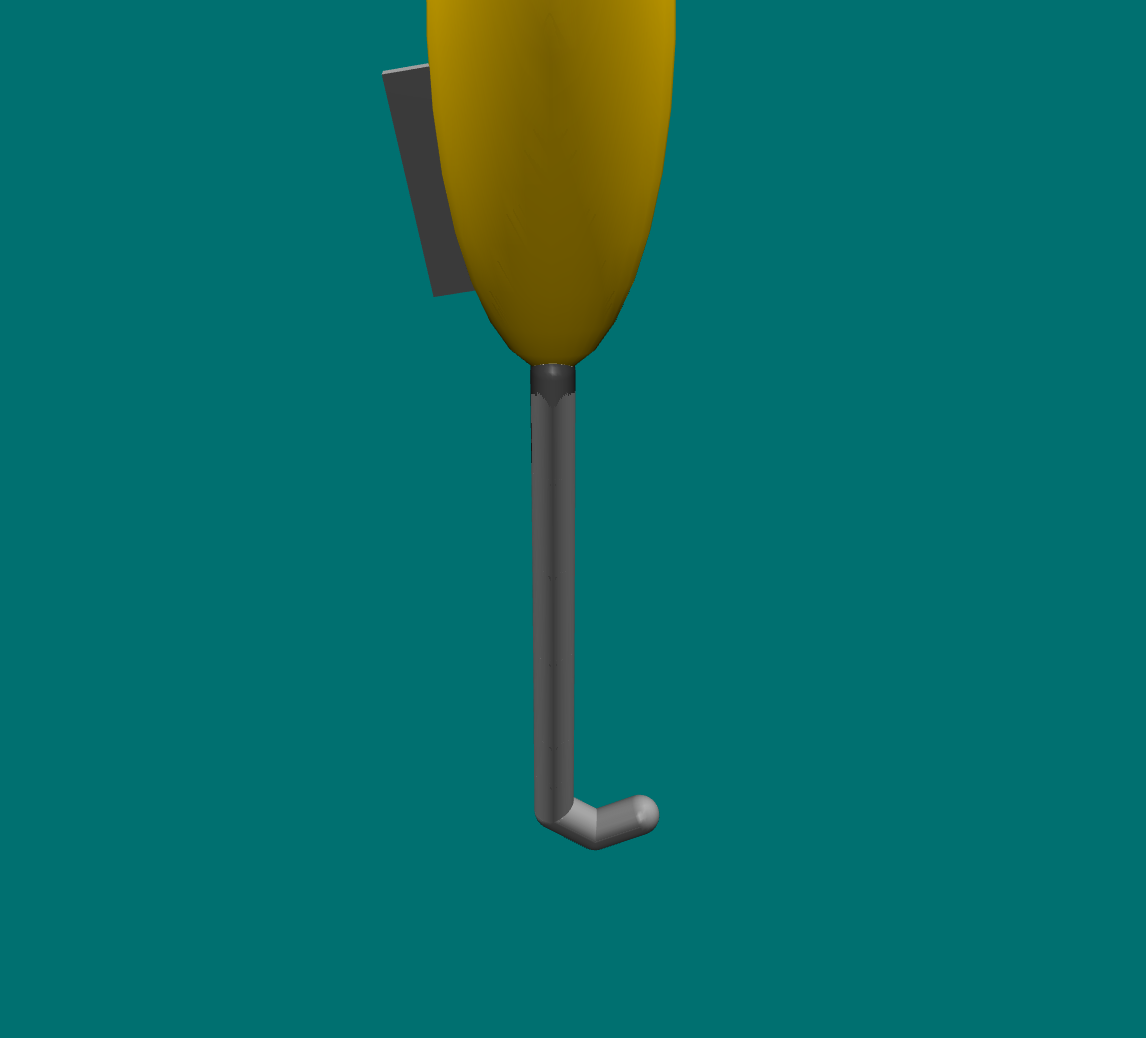} &
    \includegraphics[width=0.33\linewidth]{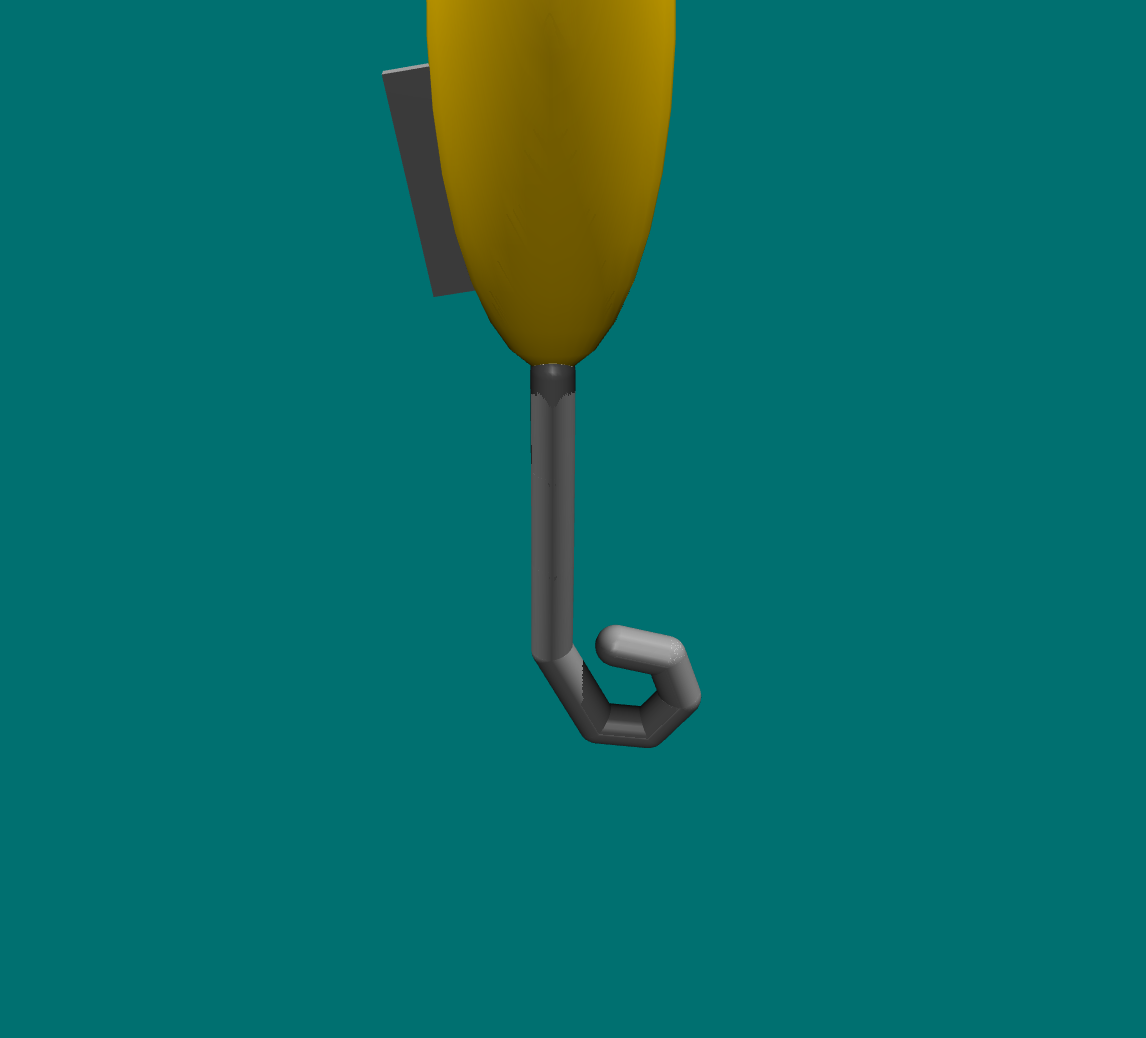}\\
    \includegraphics[width=0.33\linewidth]{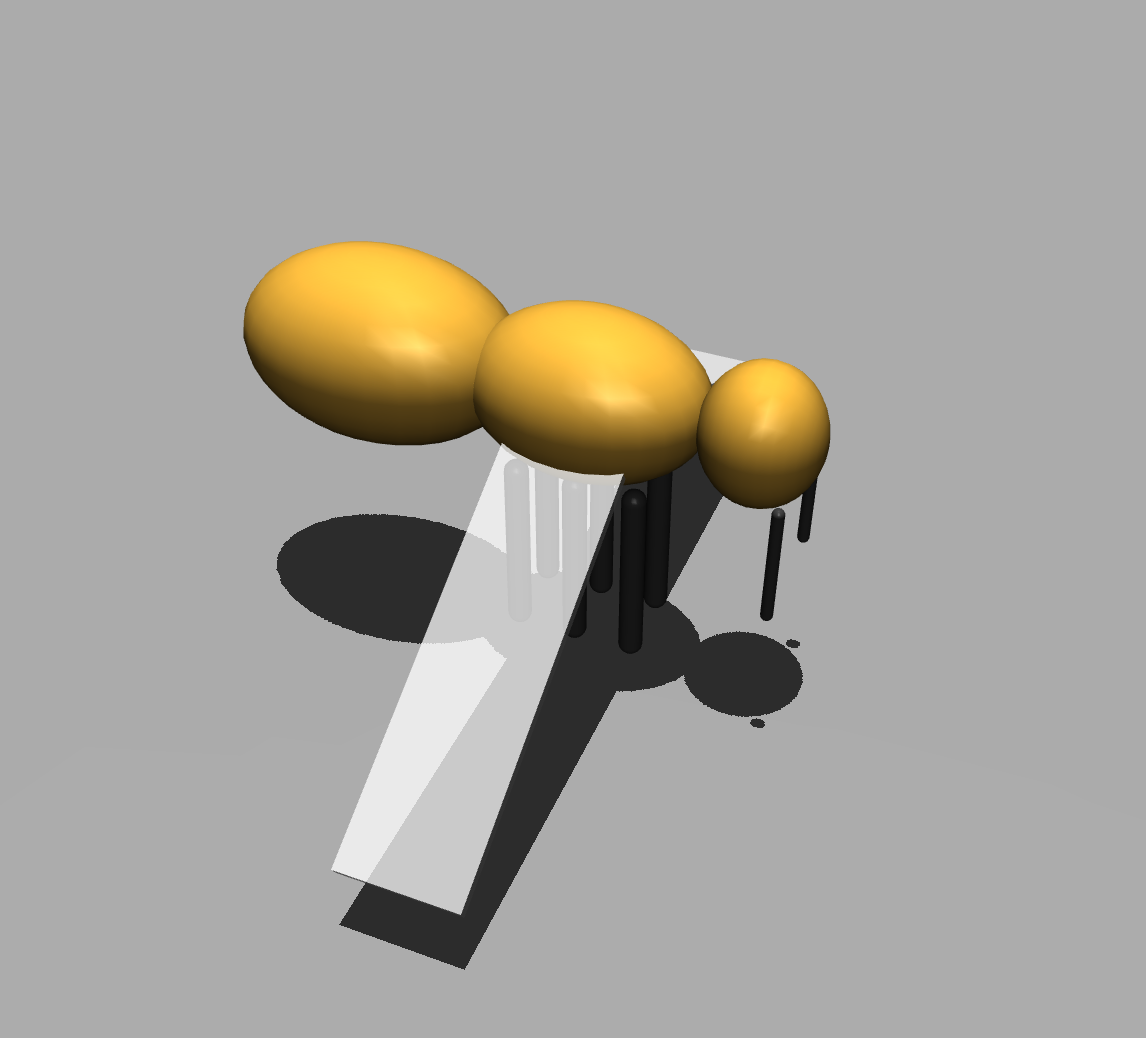} & 
    \includegraphics[width=0.33\linewidth]{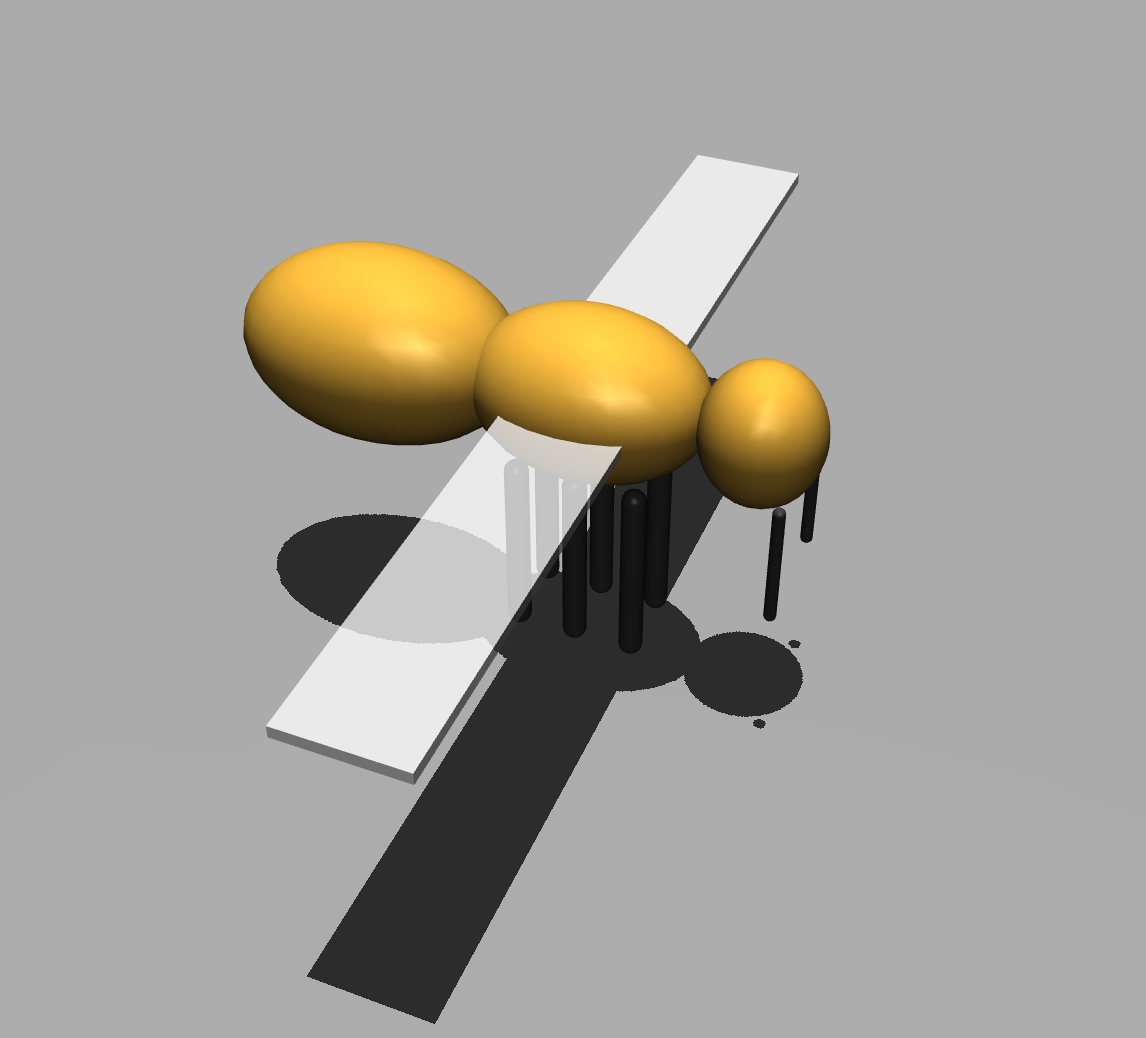} &
    \includegraphics[width=0.33\linewidth]{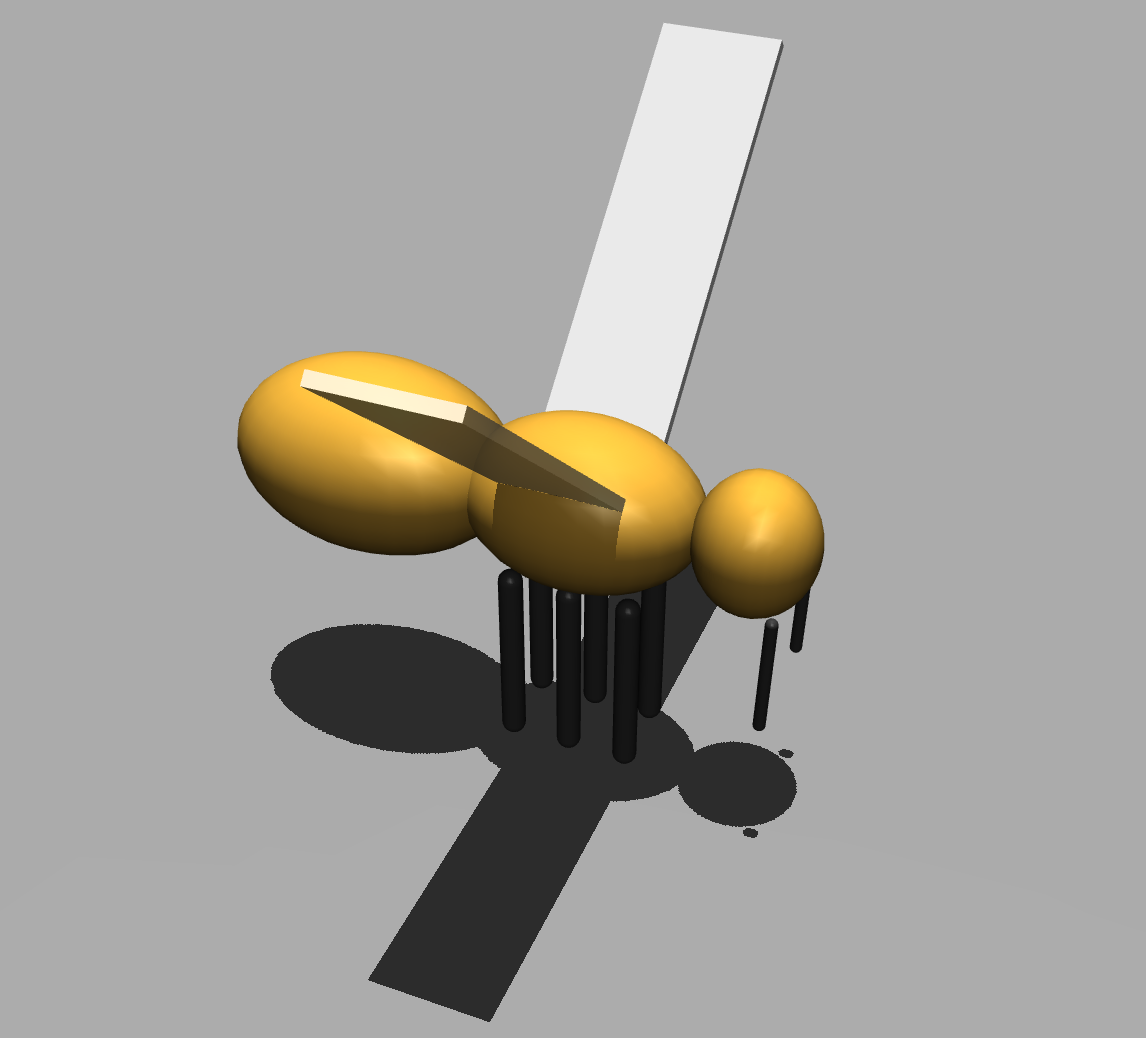}
    
    \end{tabular}
    
    \caption{Realistic joint actuation mechanisms designed by our method. The top row depicts the tail actuation mechanism of a seahorse robot while the bottom row demonstrates the wing flapping mechanism of a bee robot.}
    \label{fig:actuation_illustration}	
\end{figure}

\noindent \textbf{Joint structure.} Actuation is an important aspect of robot design that defines the space of motions the robot can perform. 
Using the general purpose reasoning capacity of VLMs, our approach selects joint types that accurately model the motions that a robot can perform. 
Our framework successfully reasons about various joint types, including hinge joints (leg joints of bipedal and quadrupedal animals), ball joints (trunk of the elephant), and fixed joints (horns of the addax), as described in Section~\ref{sec:build}. Fig.~\ref{fig:actuation_illustration} illustrates a couple of examples, including the tail actuation mechanism of a seahorse (Top) and the hinge mechanism of bee wings (Bottom).

\begin{figure}[h]
    \centering
    \setlength{\tabcolsep}{1pt}
    \renewcommand{\arraystretch}{0.7}
    \begin{tabular}{c c c}
    \includegraphics[width=0.155\textwidth]{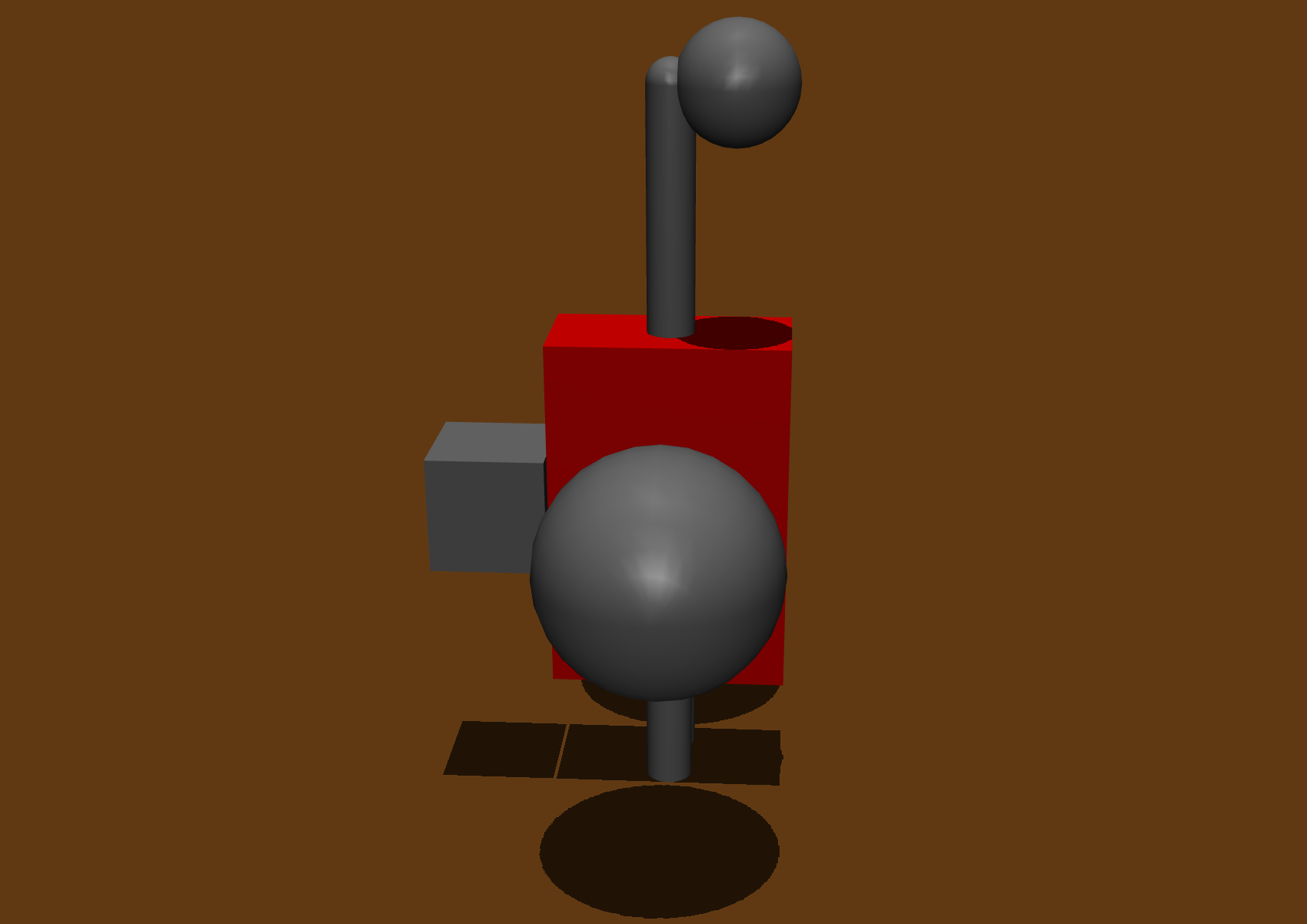} &
    \includegraphics[width=0.155\textwidth]{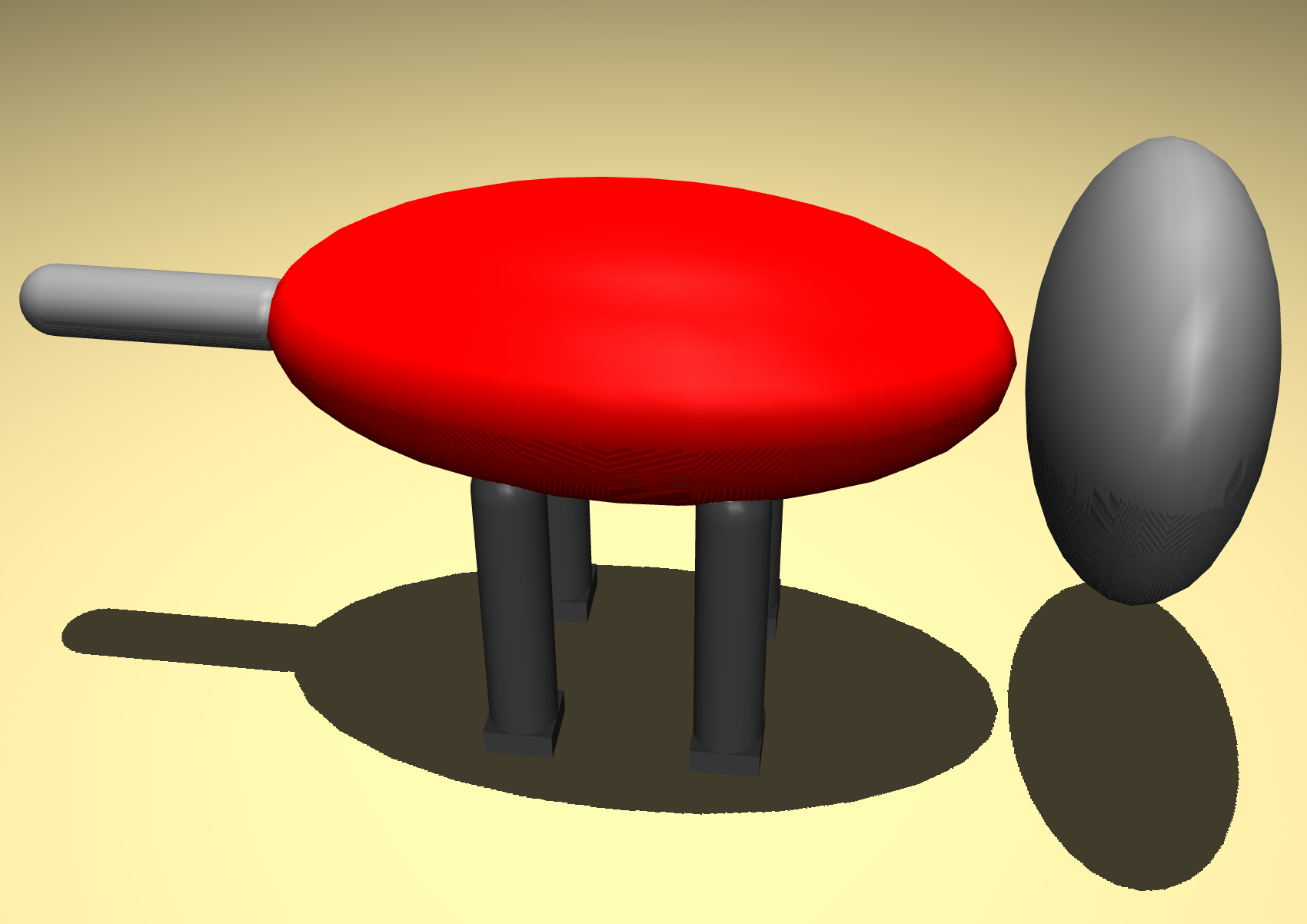} &
    \includegraphics[width=0.155\textwidth]{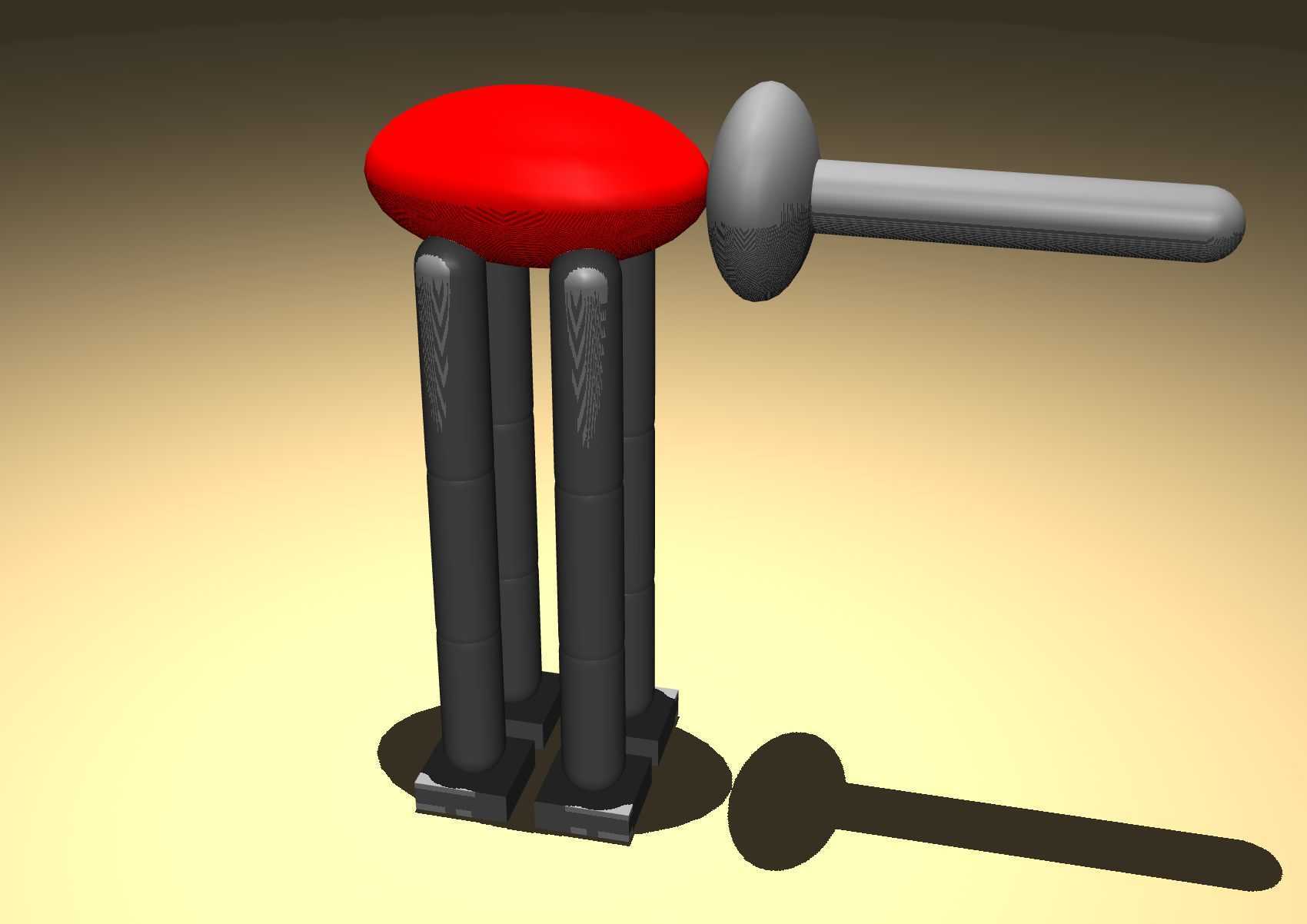} \\
    \includegraphics[width=0.155\textwidth]{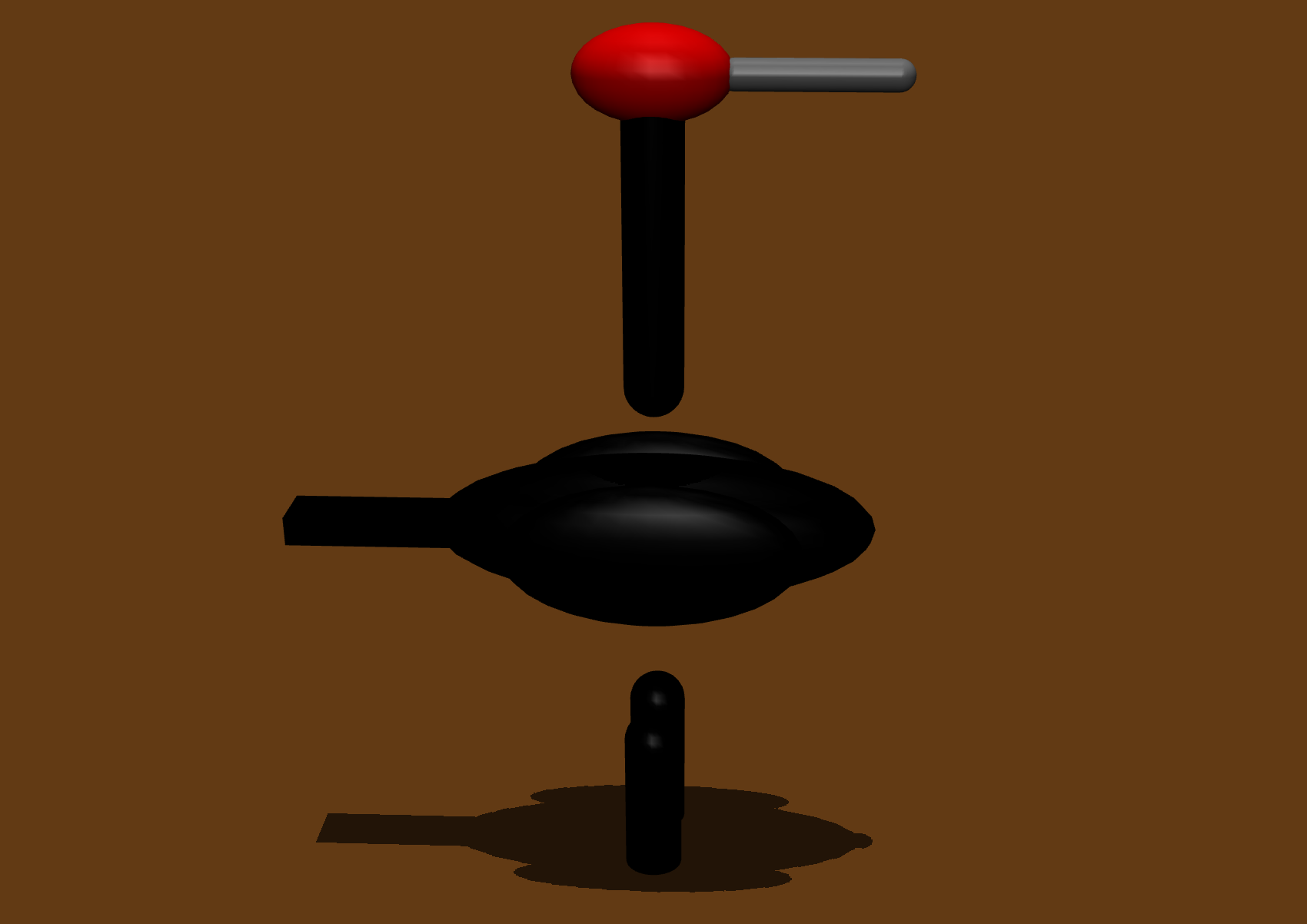} &
    \includegraphics[width=0.155\textwidth]{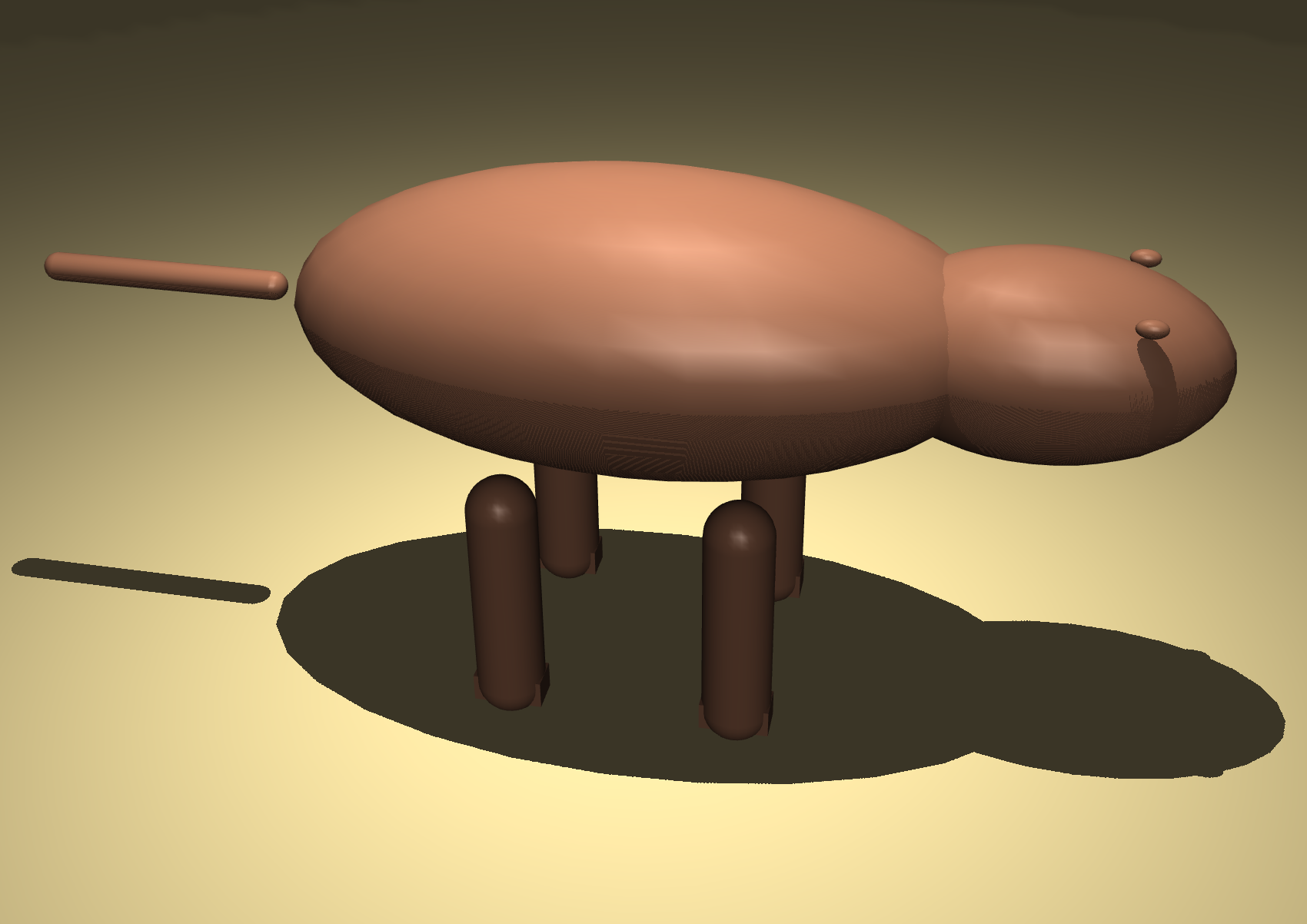} &
    \includegraphics[width=0.155\textwidth]{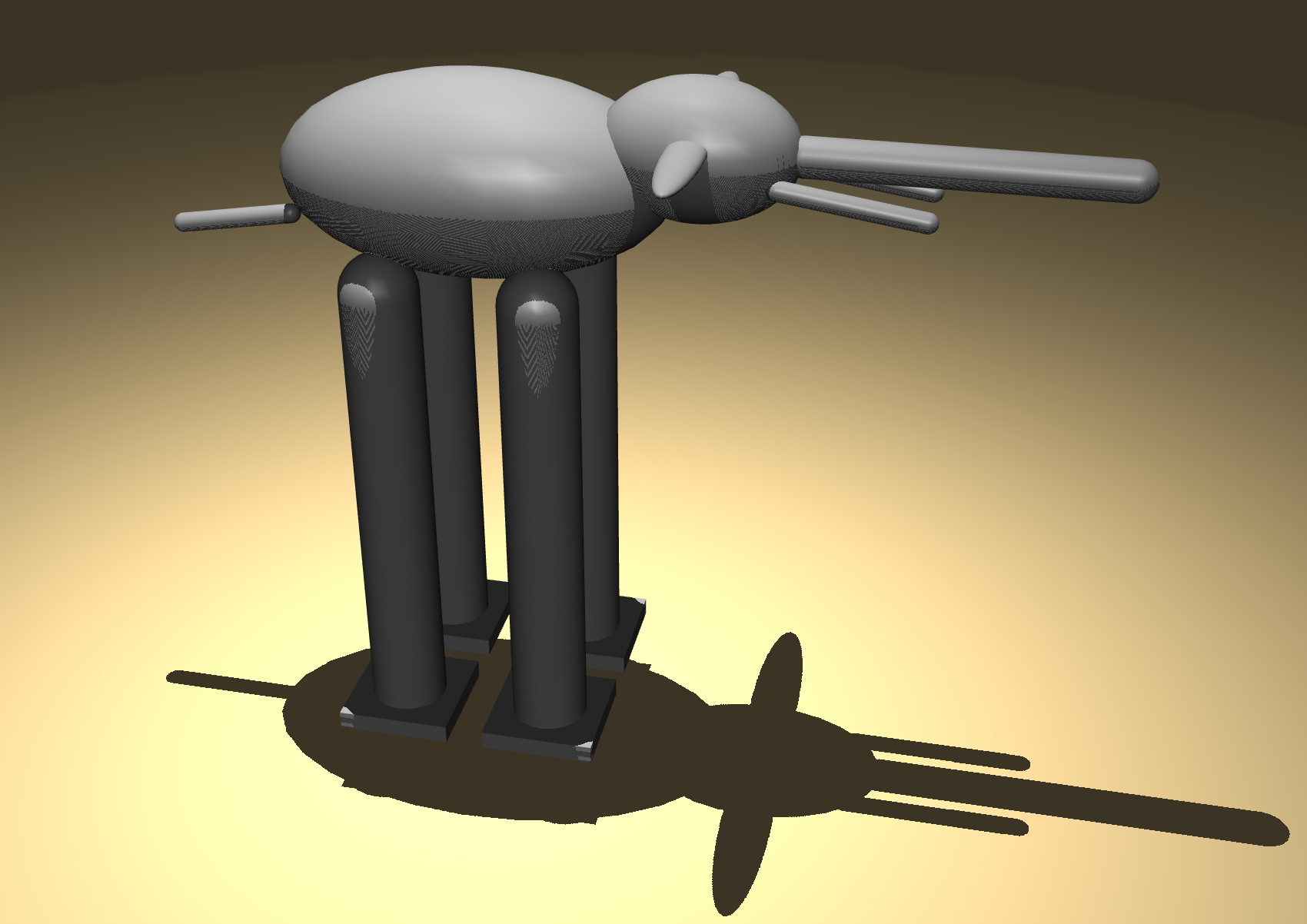} \\
    \includegraphics[width=0.155\textwidth]{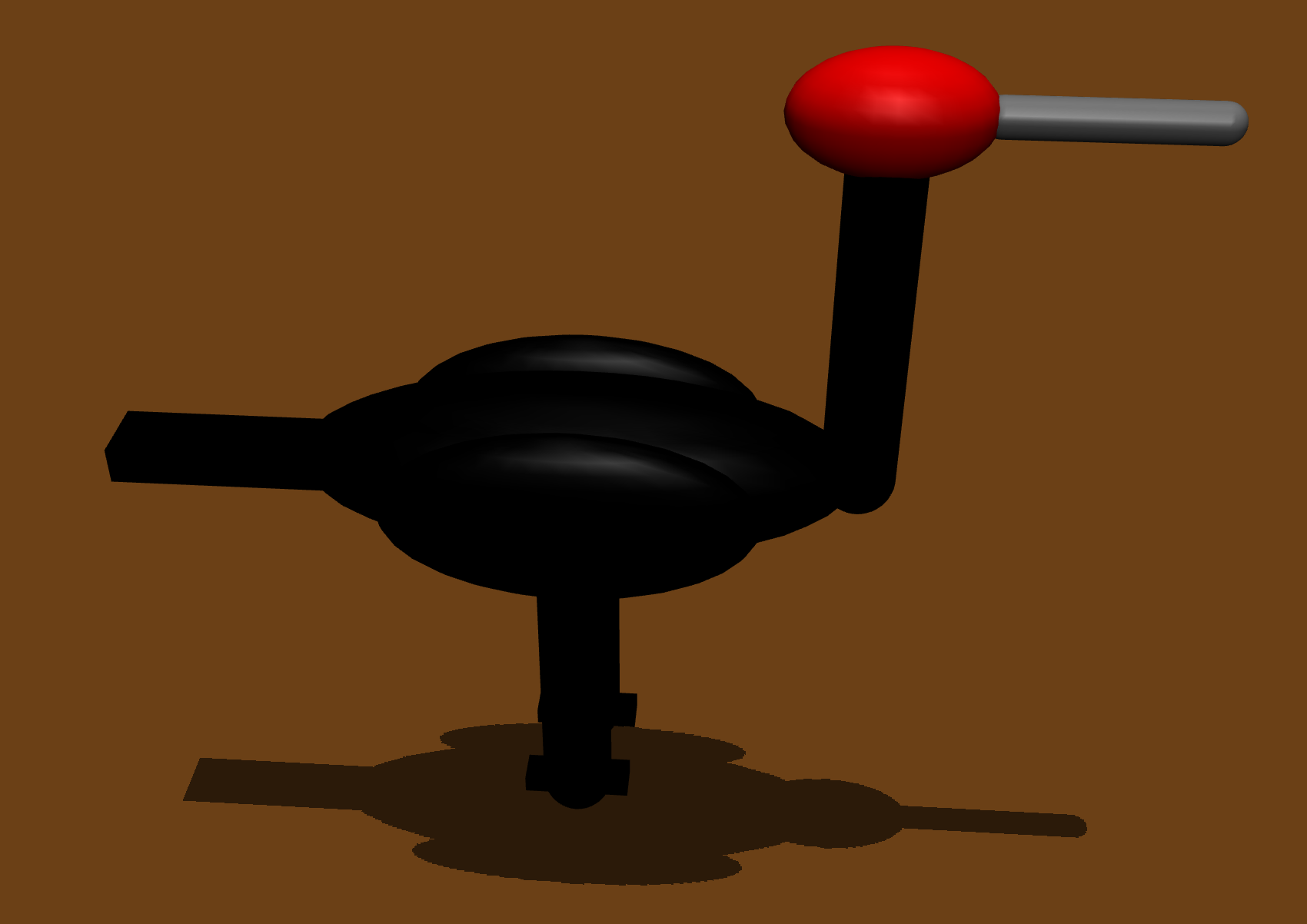} &
    \includegraphics[width=0.155\textwidth]{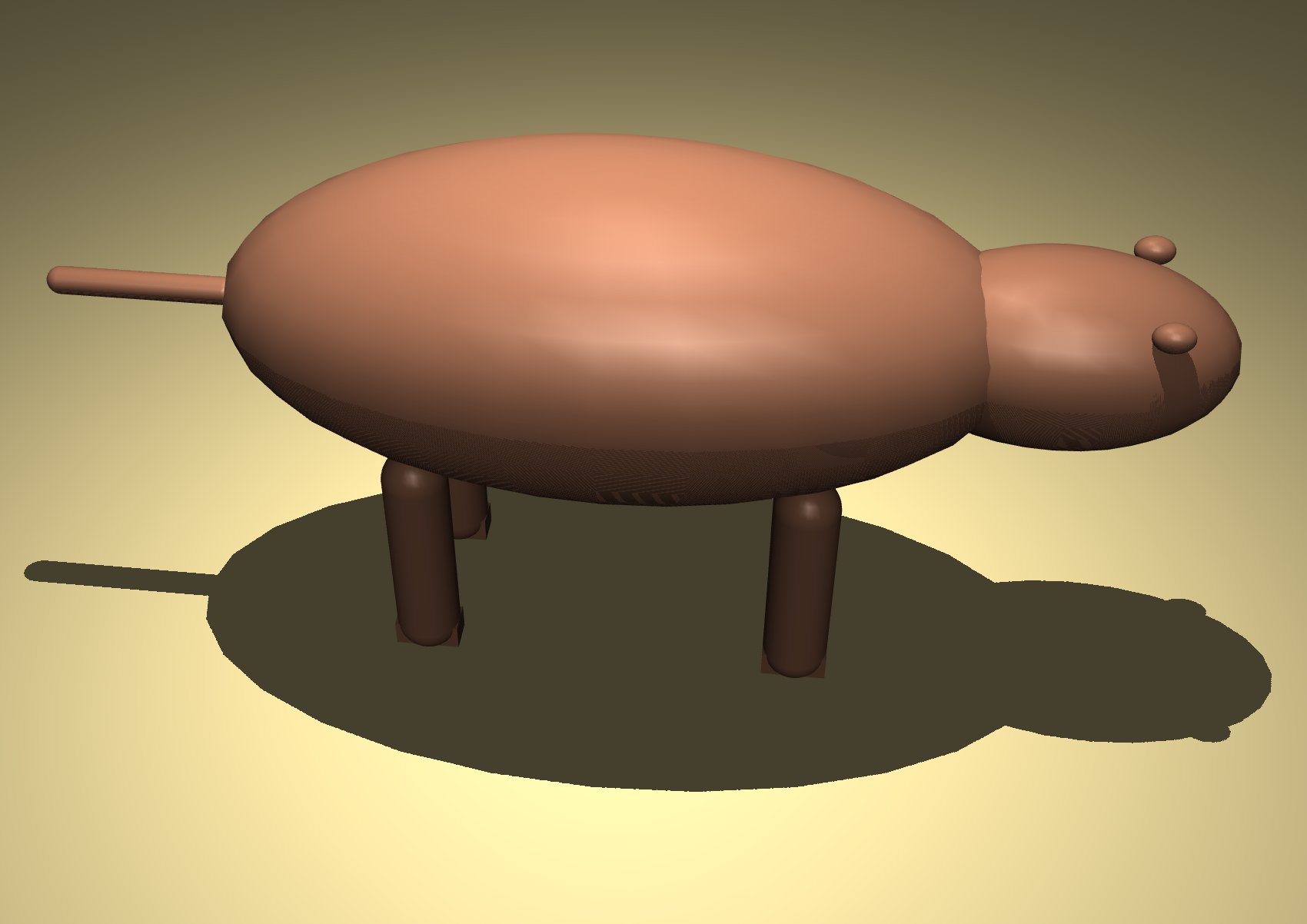} &
    \includegraphics[width=0.155\textwidth]{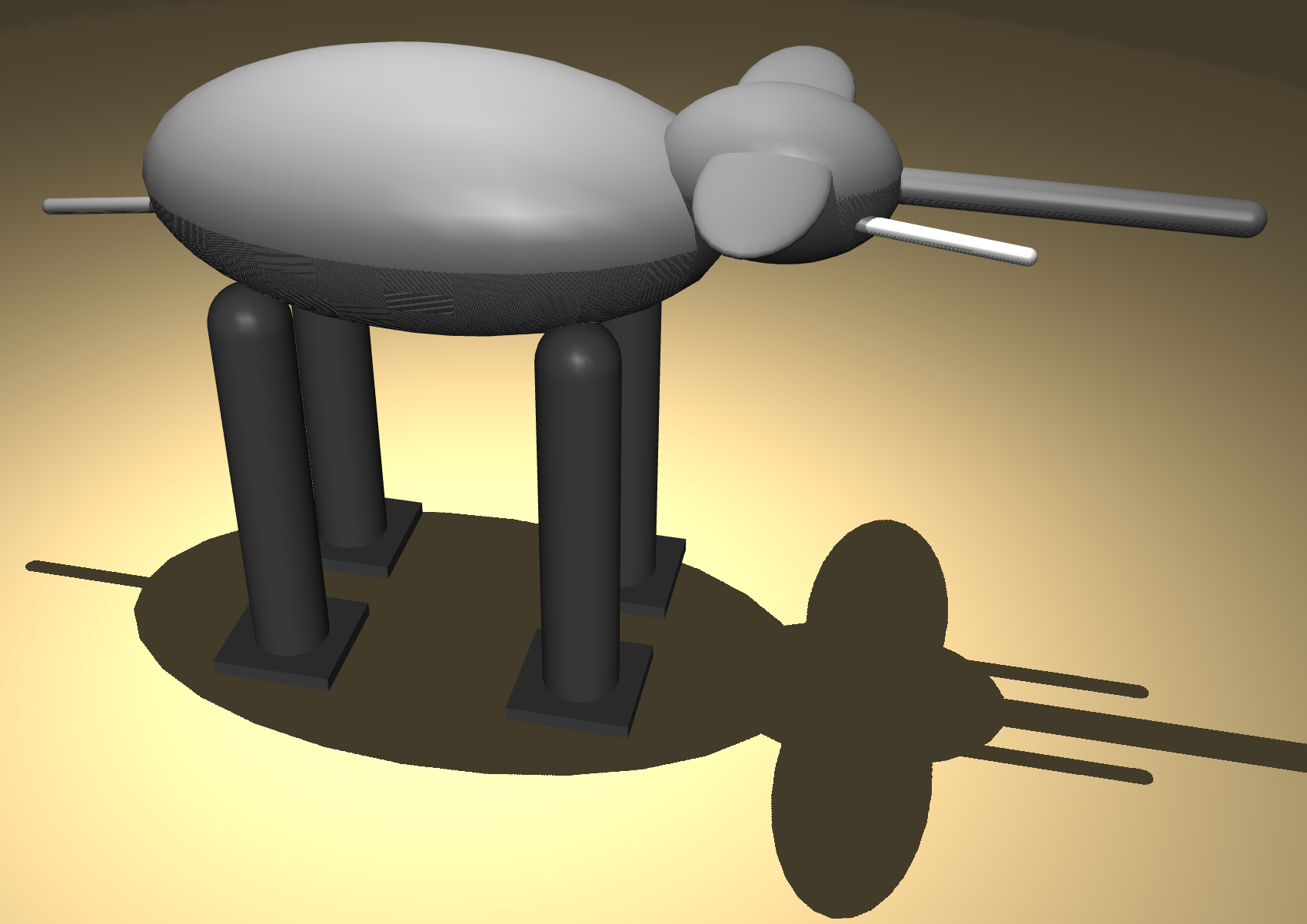} \\
    \includegraphics[width=0.155\textwidth]{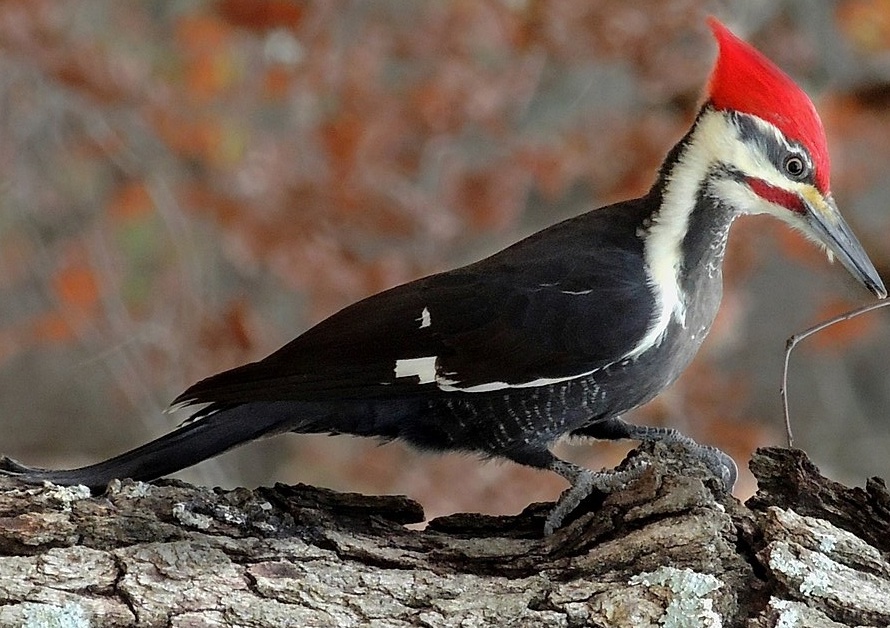} &
    \includegraphics[width=0.155\textwidth]{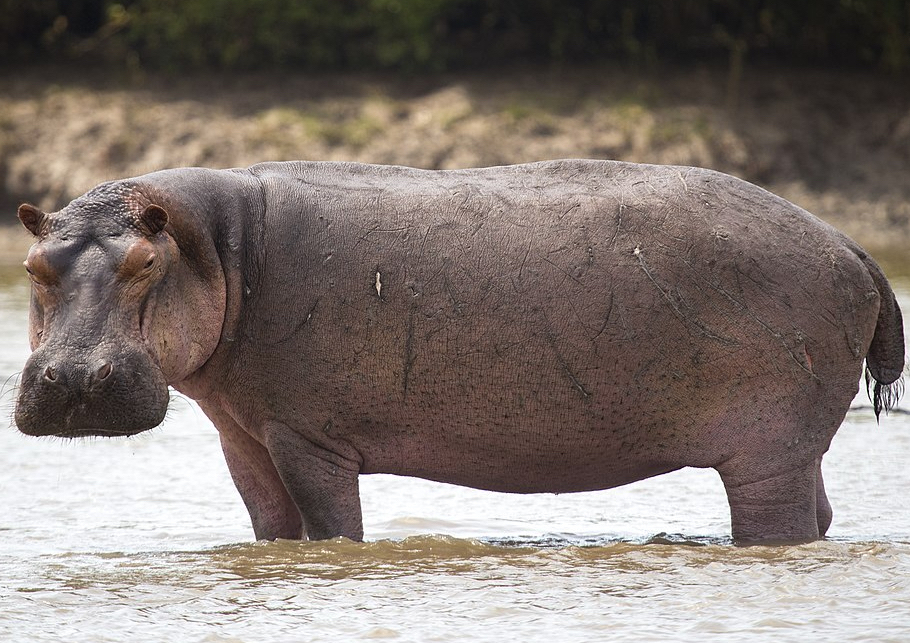} &
    \includegraphics[width=0.155\textwidth]{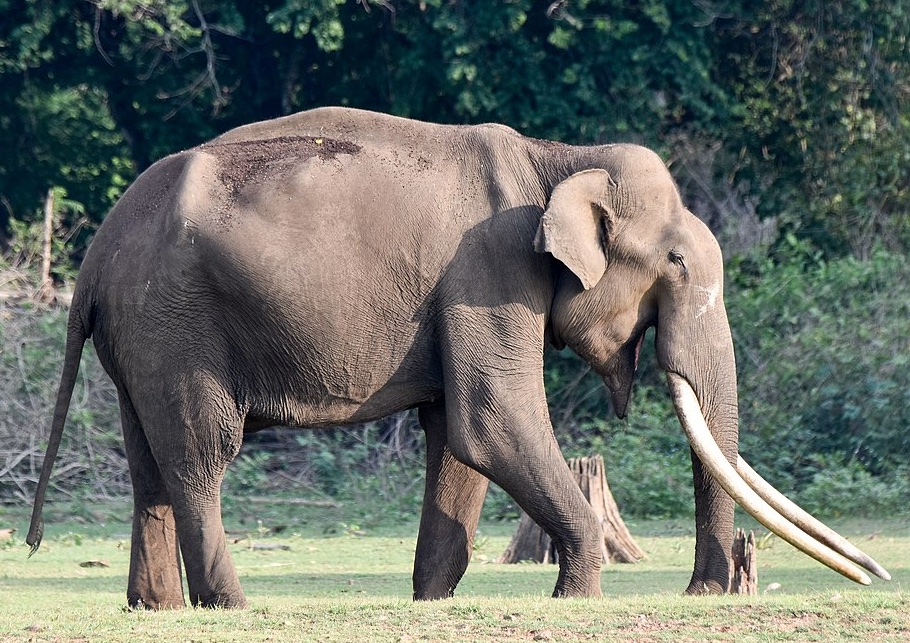} \\
    \end{tabular}
    
    \caption{Ablations. Each column shows one animal (Woodpecker, Hippopotamus, Elephant), and each row shows a condition: initial design, visual feedback, human feedback, and reference (from top to bottom).}
    \label{fig:ablation}	
\end{figure}

    

\noindent\textbf{Design details.} Our framework captures many details, such as colors, transparency, and small props, which may not be necessary for functional movements but are important for visual appearance. For instance, our framework successfully captures the transparency in the wings of the bee and the dragonfly, the dots on the back of the ladybug, the tail tuft of the bison, the two horns of the rhinoceros, the various fins of the seahorse, the ears of the hippopotamus, the hump on the camel, and the elephant's tusks.


\subsection{Ablations}
To better understand the impact of the different stages of our method on the final result, we perform an ablation study. We compare the results obtained after the text-only incremental build stage, the visual feedback stage, and finally, human feedback. We present three such examples in Fig.~\ref{fig:ablation}. Additional examples are included in the supplementary video. We observe that there is a significant increase in design quality from the text-only stage to the visual feedback stage. Also, human feedback provides a useful interface to fix minor visual details and structural mistakes.

\noindent \textbf{Text-only stage.} We observe that VLMs generated realistic kinematic structures for all $20$ cases, as illustrated by the example of the rabbit robot in Fig. \ref{fig:rabbit_tree}, but the relative scales of body components are not always captured accurately. Text-generated positions or orientations of components may not always be anatomically correct, and designs typically have a red torso with gray for other components.

\noindent\textbf{Visual feedback stage.} The visual feedback stage rectifies scale discrepancies between components and correctly changes colors to match reference images, as illustrated in the second column of Fig. \ref{fig:ablation}. Note that we use only a single color for each component and do not allow finer details in visualization, such as shading. Although our method does not explicitly add components like eyes, ears, or tusks in the text-only stage, we notice that visual feedback results in some of these features being added to the robots,
making designs significantly more realistic.

\noindent\textbf{Human feedback stage.} 
Human feedback addresses remaining errors such as disconnected components or incorrect body positions. Additionally, end users may wish to edit the designs to suit their preferences or incorporate specific constraints. In the examples presented in Fig.~\ref{fig:ablation}, these edits include ensuring the hippopotamus's (row 2), and woodpecker's (row 3) components are connected, resizing the torsos of the elephant (row 1) and the hippopotamus (row 3), and repositioning the woodpecker's neck (row 3). In our experiments, three short human feedback sessions were typically sufficient to enhance visual quality and address minor kinematic errors. The average number of feedback iterations was $2.15$ across all cases.




\subsection{User Study}

\begin{figure}
    \centering
    \includegraphics[width=0.8\linewidth]{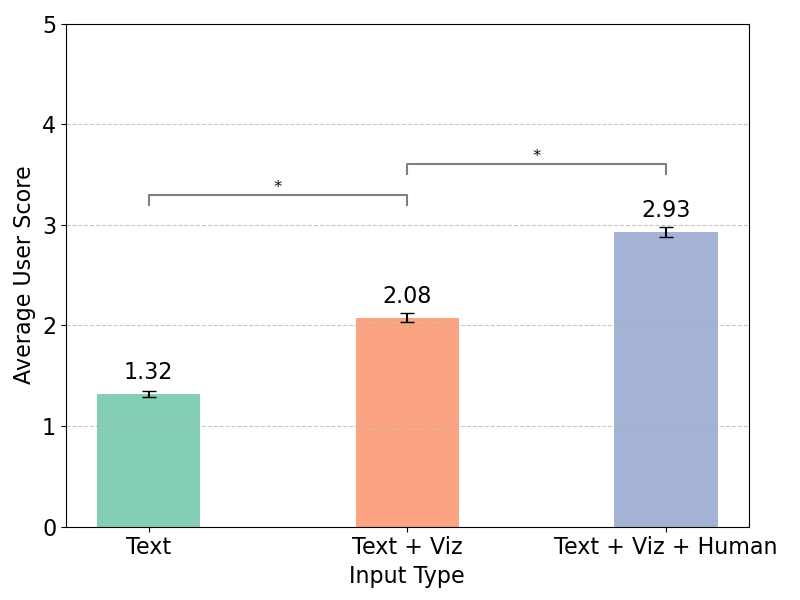}
    \caption{Results of our user survey. We plot the average user score for each of the three stages - \textit{text}, \textit{vis}, and \textit{hum}. The black bars represent the standard error of the mean.}
    \label{fig:user_survey}
\end{figure}

We performed a user study with $44$ participants to validate our results and highlight the contribution of each stage of our method. Participants are shown images depicting different animals, birds, and insects. They are then shown the outputs of each of the three stages - text-only (\textit{text}), text + visual feedback (\textit{vis}), and text + visual feedback + human feedback (\textit{hum}) - in a random order. They are asked to rate how closely each robot model resembles the real creature shown in the image. We use a subset of $10$ out of a total of $20$ designs generated by our method for this user study. Based on the data collected from this survey, we plot the graph presented in Fig.~\ref{fig:user_survey}. 
Results show that \textit{vis} is significantly better than \textit{text} (\textit{F-statistic}$=219.13, p<0.005$) and \textit{hum} is significantly better than \textit{vis} (\textit{F-statistic}$=172.28, p<0.005$). This demonstrates the notable improvements made at each step of our proposed design process.


\begin{figure*}[ht]
    \centering
    \setlength{\tabcolsep}{1pt}
    \renewcommand{\arraystretch}{0.7}
    \begin{tabular}{c c c c c}
    \includegraphics[width=0.195\textwidth]{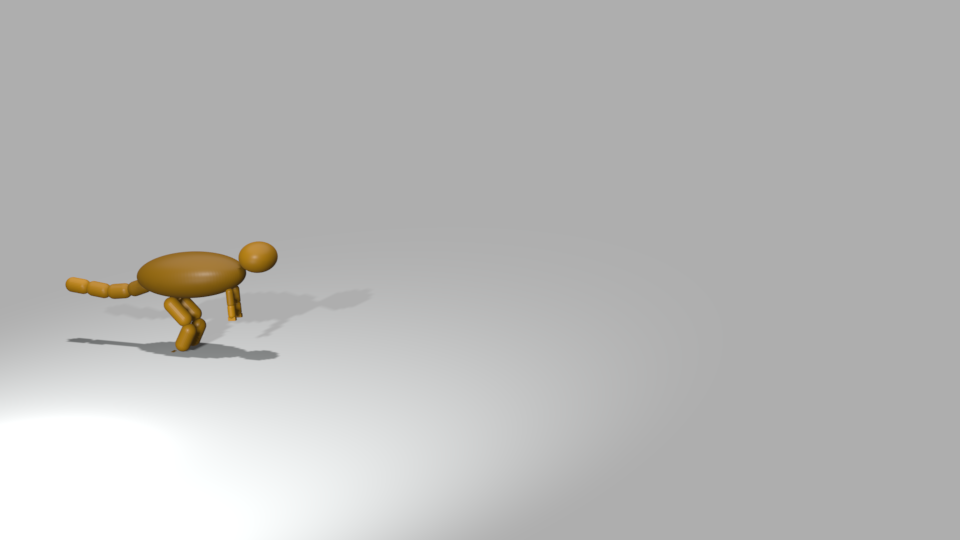} & 
    \includegraphics[width=0.195\textwidth]{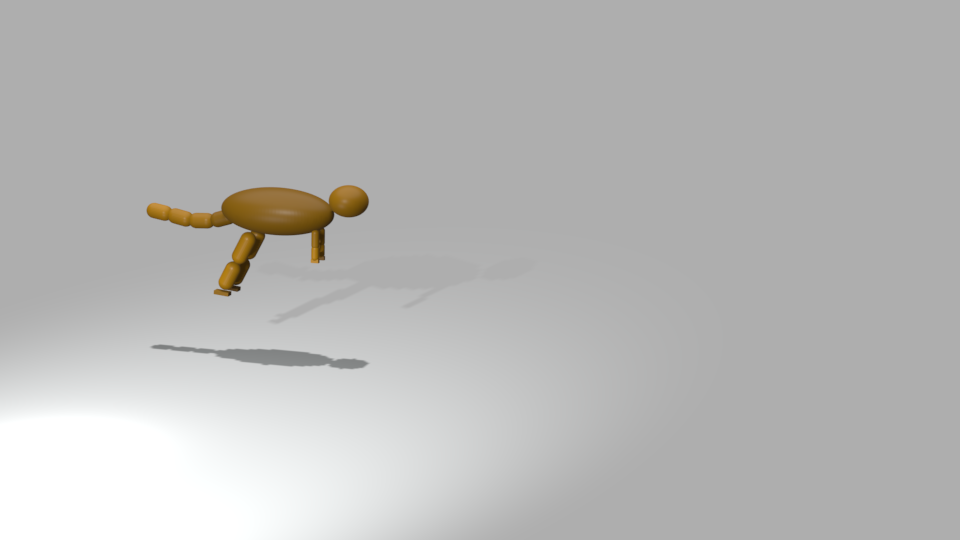} & 
    \includegraphics[width=0.195\textwidth]{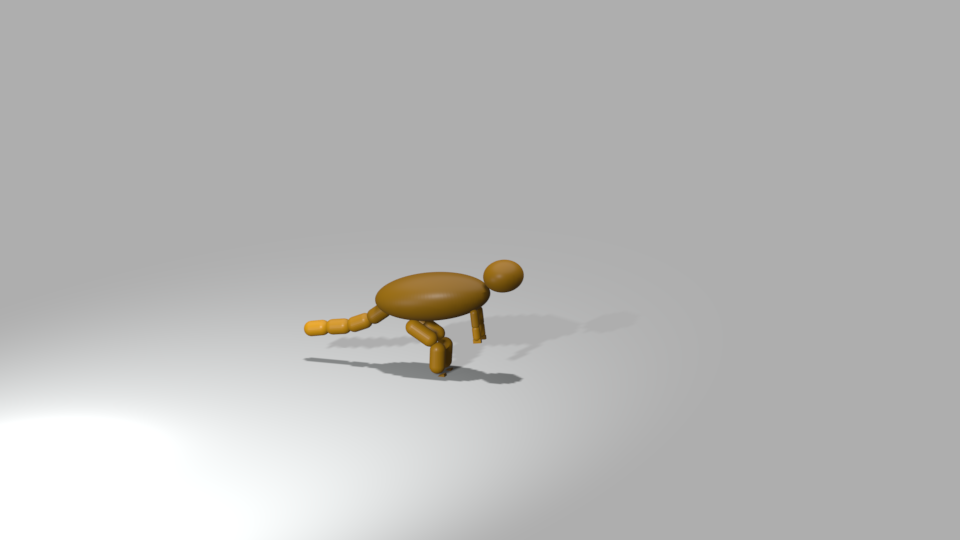} & 
    \includegraphics[width=0.195\textwidth]{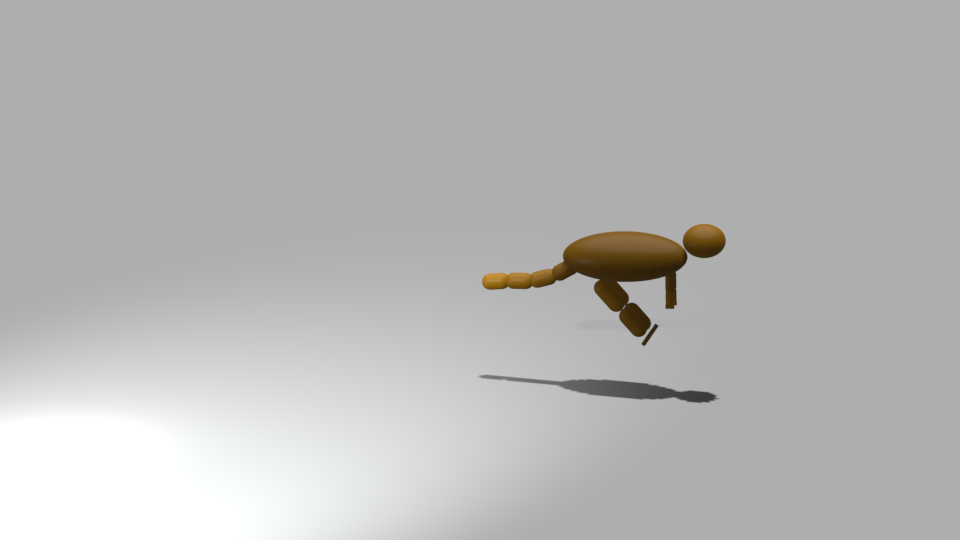} & 
    \includegraphics[width=0.195\textwidth]{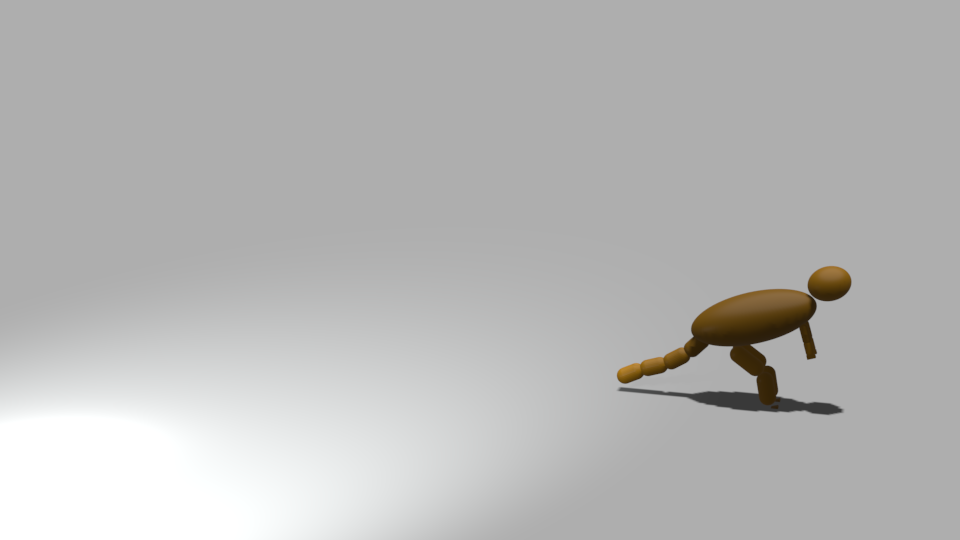} \\
    \includegraphics[width=0.195\textwidth]{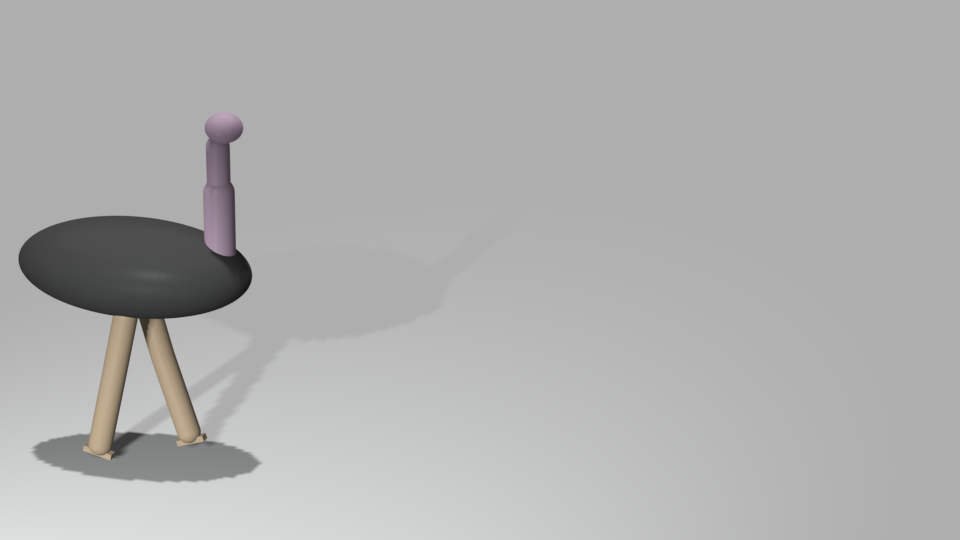} & 
    \includegraphics[width=0.195\textwidth]{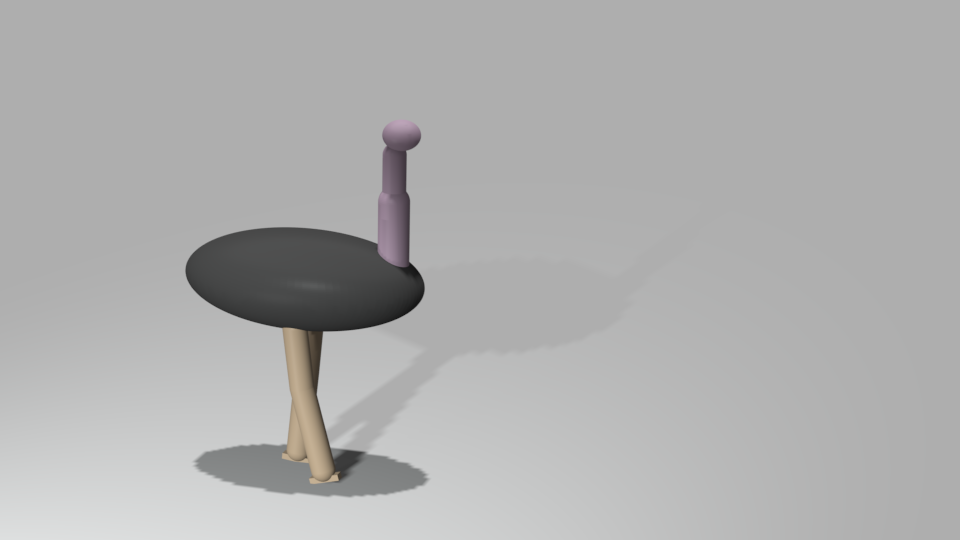} & 
    \includegraphics[width=0.195\textwidth]{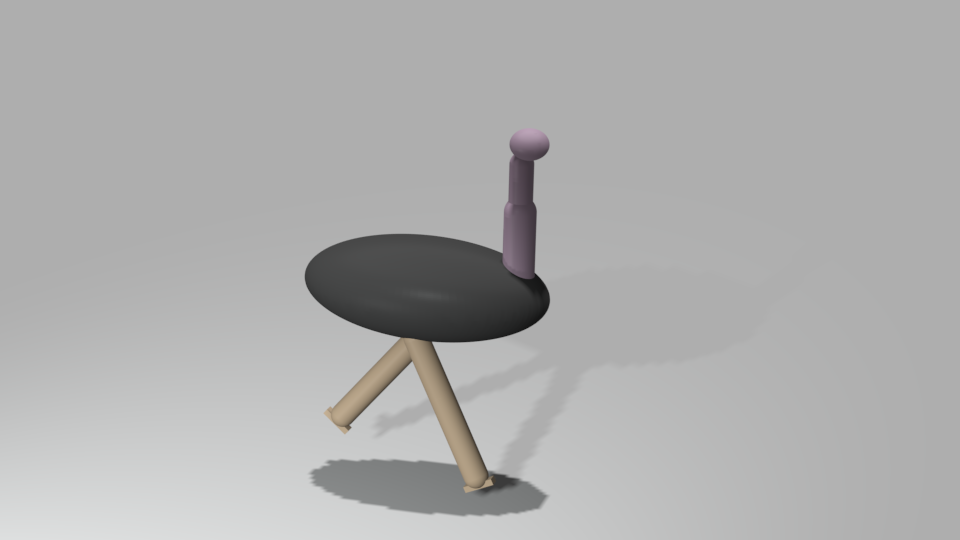} & 
    \includegraphics[width=0.195\textwidth]{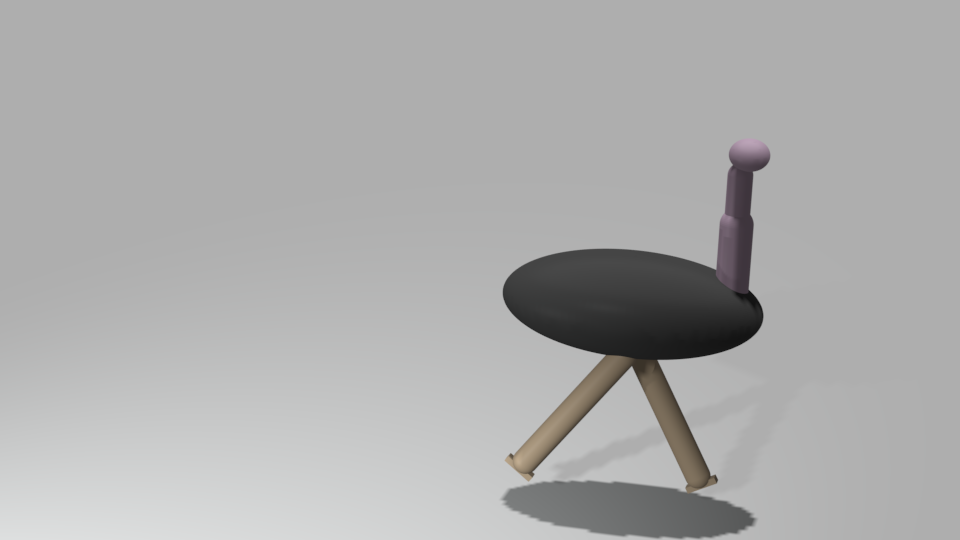} & 
    \includegraphics[width=0.195\textwidth]{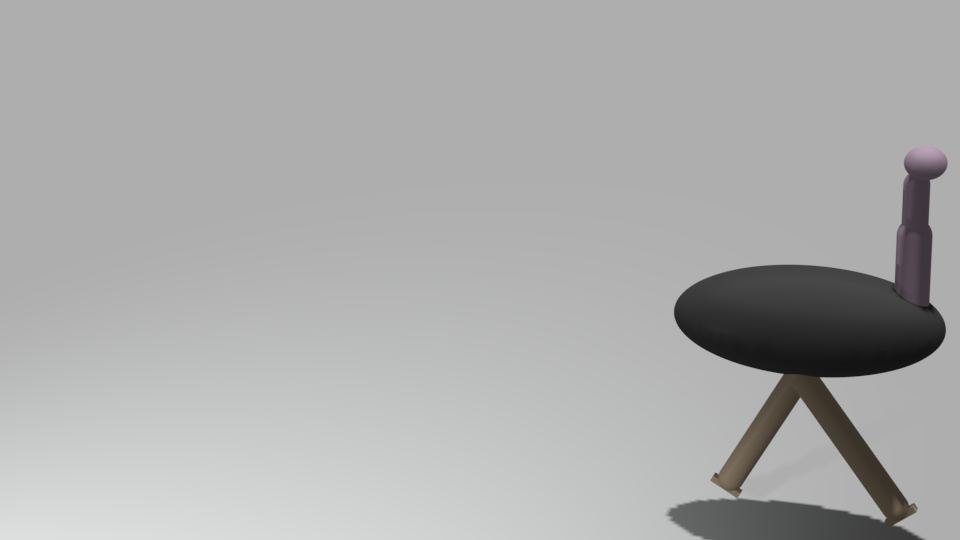} \\
    \includegraphics[width=0.195\textwidth]{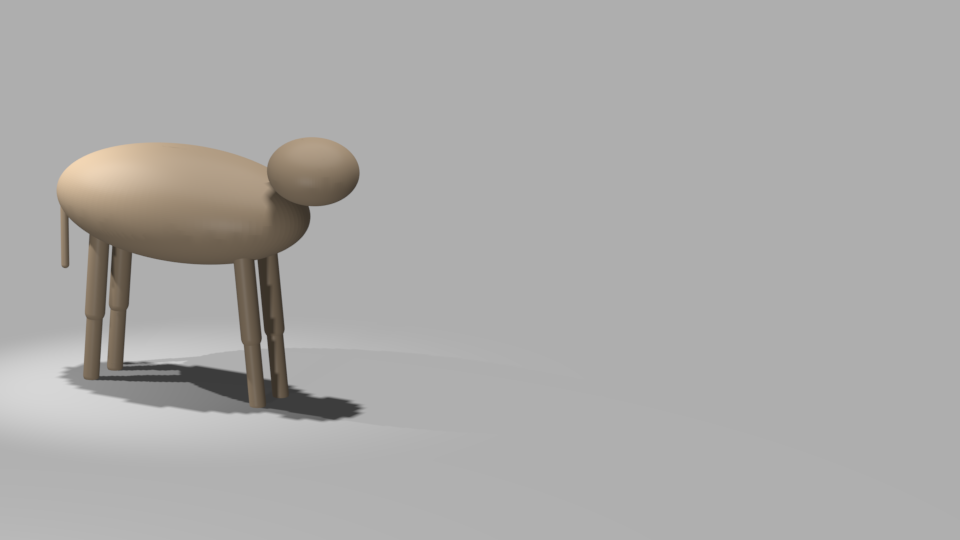} & 
    \includegraphics[width=0.195\textwidth]{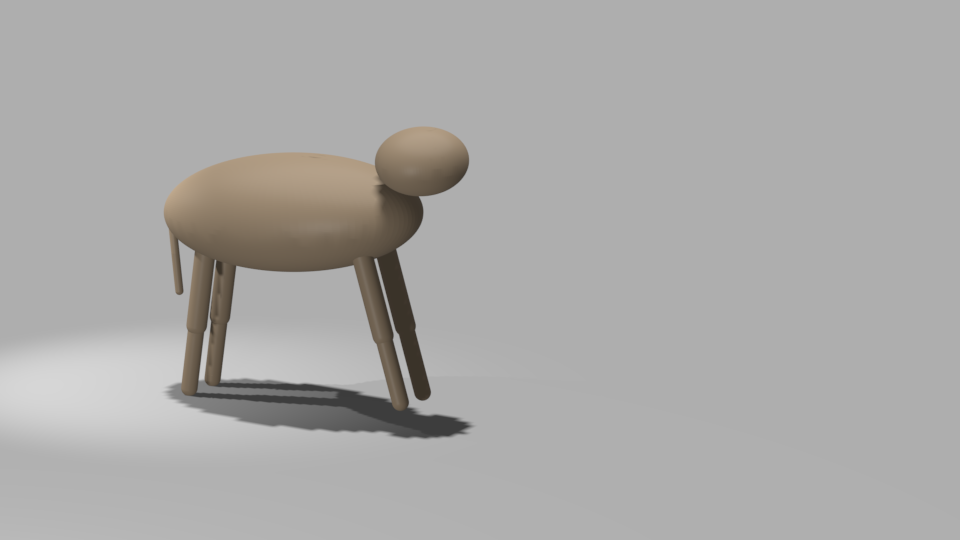} & 
    \includegraphics[width=0.195\textwidth]{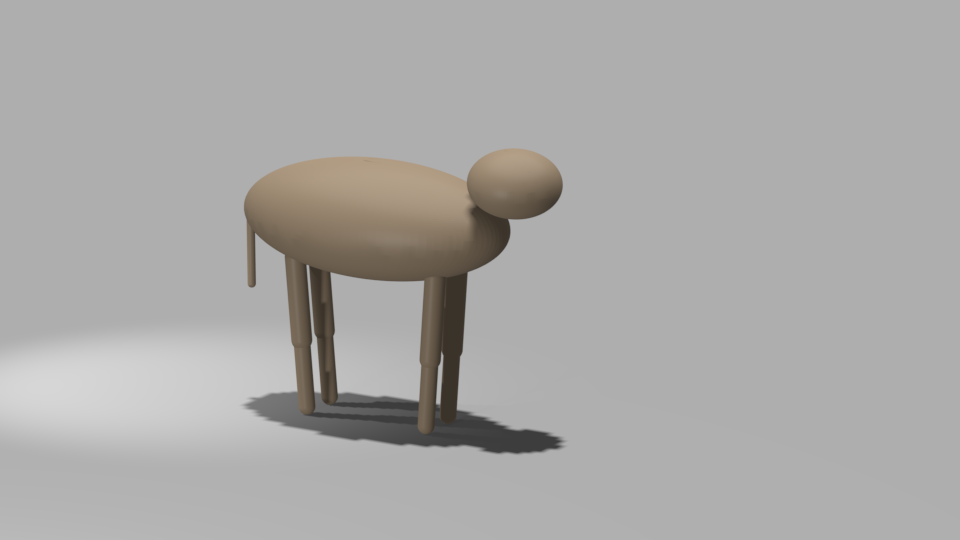} & 
    \includegraphics[width=0.195\textwidth]{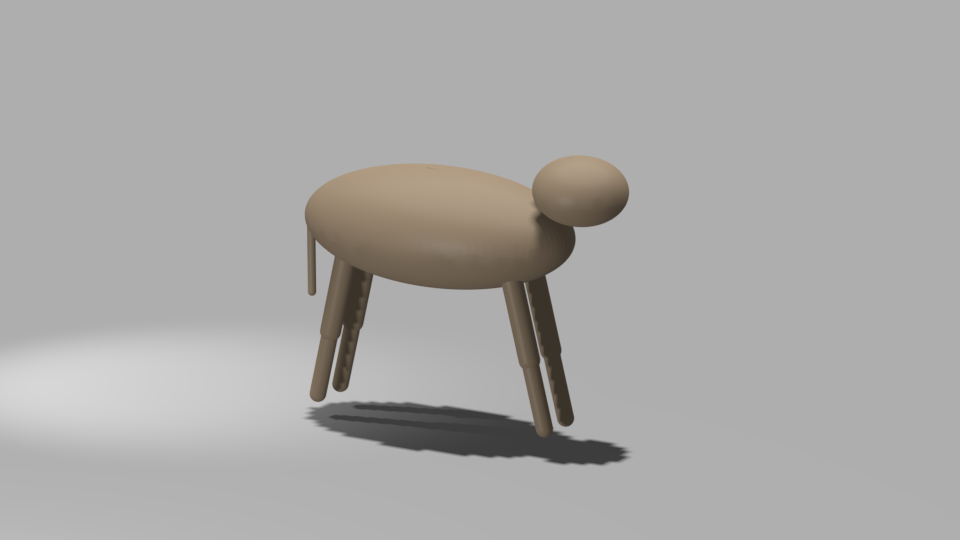} & 
    \includegraphics[width=0.195\textwidth]{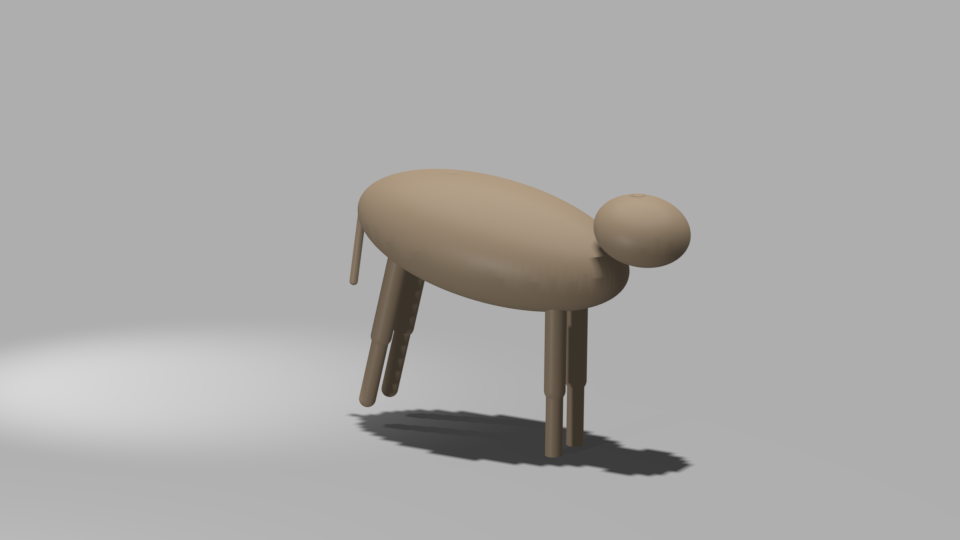} \\

    \end{tabular}
    \caption{Synthesized motions of kangaroo hop (top), ostrich run (middle), and dog bound (bottom).
    }
    \label{fig:motion_strip}	
\end{figure*}

\subsection{Motion Generation} \label{sec:motion}
To demonstrate the kinematic capabilities of the synthesized robot designs, we generate motions using an existing trajectory optimization package TOWR~\cite{winkler18}, which is designed to produce optimal motion trajectories for various legged robots, including bipeds and quadrupeds.
Fig.~\ref{fig:motion_strip} illustrates the hopping gait of the kangaroo robot, the running gait of the ostrich robot, and the bounding gait of the dog robot. The motion planning package successfully generated realistic and reasonable gaits. The objective costs were low enough to ensure the physical validity of the synthesized motions. 

\subsection{Comparison to Baselines} \label{sec:baselines}



We provide a qualitative comparison of the proposed method to LLM and VLM-based design methods in Table~\ref{tab:related work}. We compare the input modalities, types of objects being designed, output formats, and the degree of human involvement. 
We want to stress that our method is only the framework that can synthesize arbitrary articulated robots, while many other methods assume specific morphologies, such as quadrupeds~\cite{ringel2024text2robot}, quadcopter~\cite{makatura2023can,makatura2024can}, soft robots~\cite{ma2024exploration}, and modular robots~\cite{qiu2024robomorph}.
Of the methods listed in Table~\ref{tab:related work}, we find that Articulate-Anything~\cite{le2024articulate} and Text2Robot~\cite{ringel2024text2robot} are the closest points of comparison.

Articulate-Anything's (AA) asset generation differs from ours in two central ways. First, its mesh generation process is primarily retrieval-centric, matching user queries or visual descriptors to parts in large existing databases. While AA demonstrates mesh generation from images using newer 3D generative models, its canonical pipeline optimizes for composition, placement, and articulation of ``known'' geometries rather than unconstrained synthesis from prompts. By contrast, our framework can originate novel geometric and kinematic structures from scratch. Second, while Articulate-Anything demonstrates impressive open-ended articulated object modeling, its success hinges on asset-rich datasets containing thousands or millions of annotated human artifacts. In contrast, there are far fewer publicly available, richly-annotated datasets for robot models, making direct transfer of their strategies for link placement and affordance prediction to the robotics domain difficult. This data scarcity highlights the strength of our method, which does not rely on large training corpora or priors over robot forms and can generate plausible robot topologies directly from unconstrained prompts.


\begin{table}
    \centering
    \begin{tabular}{|>{\centering\arraybackslash}m{4.5em}|>{\centering\arraybackslash}m{4.5em}|>{\centering\arraybackslash}m{4.5em}|>{\centering\arraybackslash}m{4.5em}|>{\centering\arraybackslash}m{4.5em}|}
        \hline
        \textbf{Method} & \textbf{Asset Type} & \textbf{Input Modality} & \textbf{Process} & \textbf{Output Type} \\
        \hline
            Text2Robot \cite{ringel2024text2robot} & Quadrupedal robots & Text-only & Human-in-the-loop & URDF \cite{urdf_wiki} robot model \\
        \hline
            Articulate-Anything \cite{le2024articulate} & Articulated 3D objects & Text + vision & Automated & URDF~\cite{urdf_wiki} model\\
        \hline
            Stella~\etal \cite{stella2023can} & Robot grippers & Text-only & Human-in-the-loop & High-level specs \\
        \hline
            Makatura \etal \cite{makatura2023can, makatura2024can} & Quadcopters & Text-only & Human-in-the-loop & URDF \cite{urdf_wiki} robot model \\
        \hline
            Ma~\etal \cite{ma2024exploration} & Soft Robots & Text-only & Automated &  Natural language robot description \\
        \hline
            RoboMorph \cite{qiu2024robomorph} & Modular robots & Text-only & Automated & MJCF \cite{todorov2012mujoco} robot model \\
        \hline
            Ours & \textbf{Arbitrary robots} & Text + vision & Automated, human-editable & MJCF \cite{todorov2012mujoco} robot model \\
        \hline
    \end{tabular}
    \caption{Qualitative Comparison of LLM and VLM-based design methods to our RobotDesignGPT. Ours can synthesize arbitrary articulated robots without the limitation of types.}
    \label{tab:related work}
\end{table}

\begin{figure}
    \centering
    \setlength{\tabcolsep}{1pt}
    \renewcommand{\arraystretch}{0.7}
    \begin{tabular}{c c c c}
    \includegraphics[width=0.12\textwidth]{figures/human_feedback/seahorse.png} &
    \includegraphics[width=0.12\textwidth]{figures/human_feedback/kingfisher.png} &
    \includegraphics[width=0.12\textwidth]{figures/human_feedback/platypus.png} &
    \includegraphics[width=0.12\textwidth]{figures/human_feedback/crab.png} \\
    \includegraphics[width=0.12\textwidth]{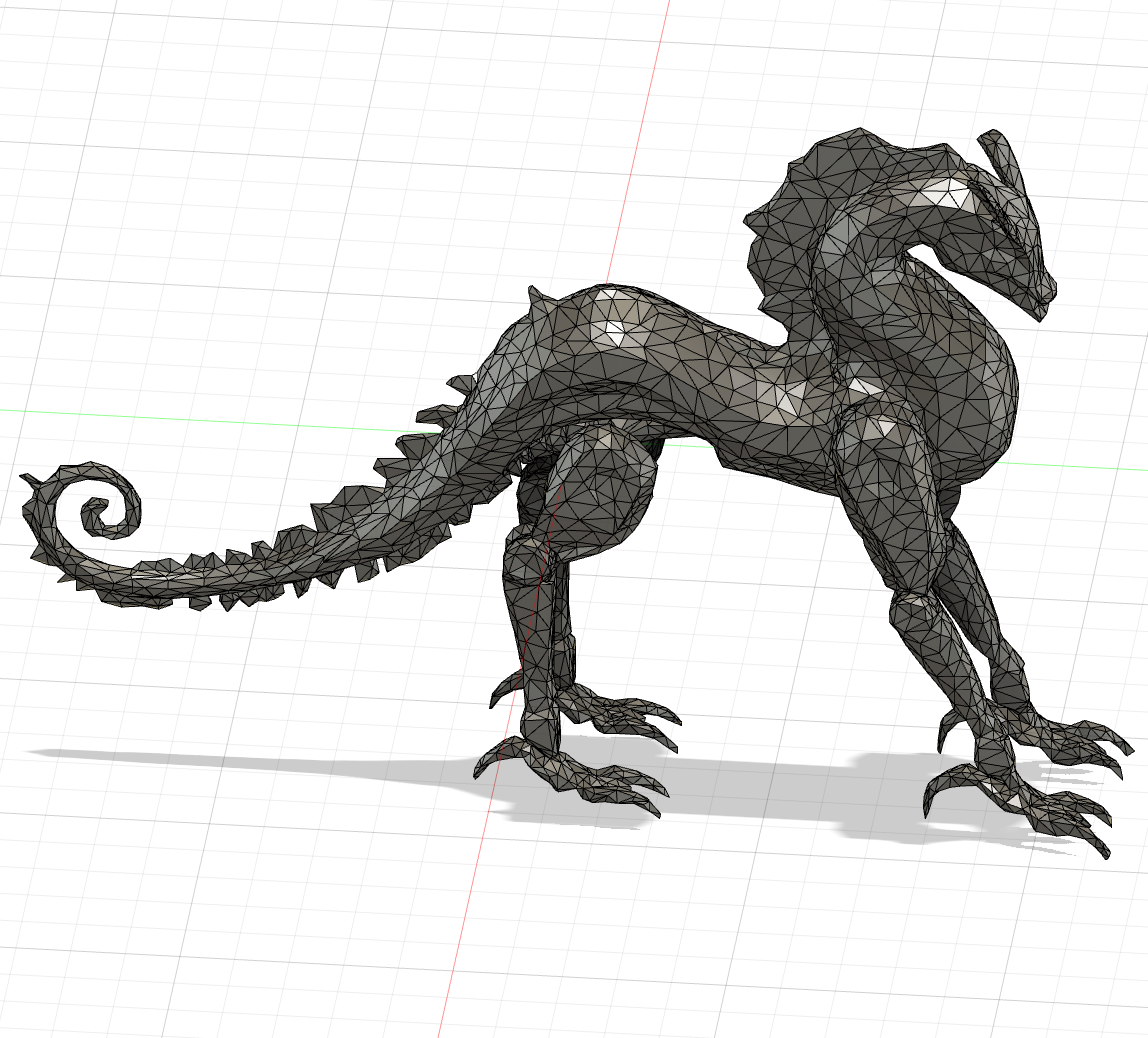} &
    \includegraphics[width=0.12\textwidth]{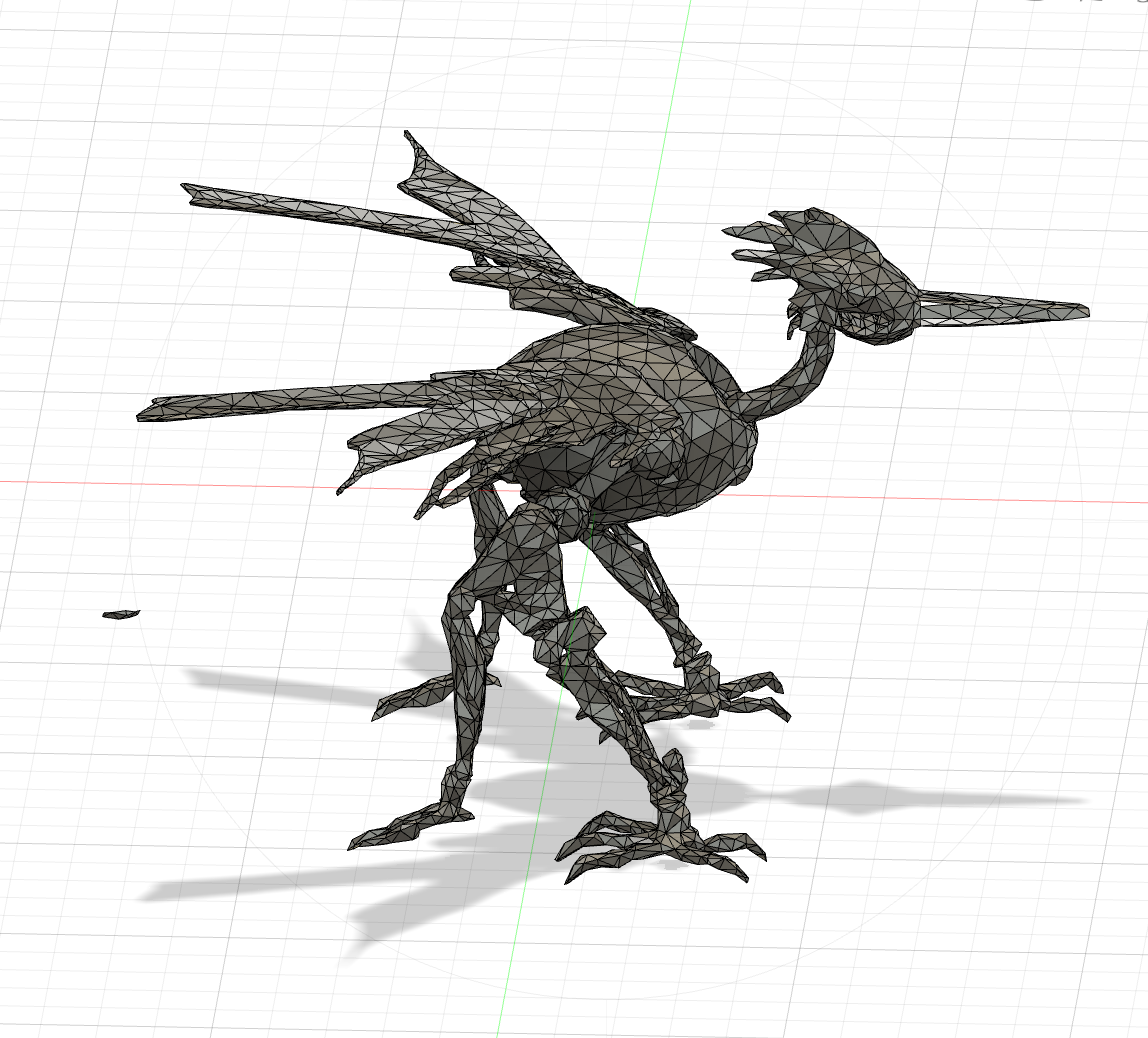} &
    \includegraphics[width=0.12\textwidth]{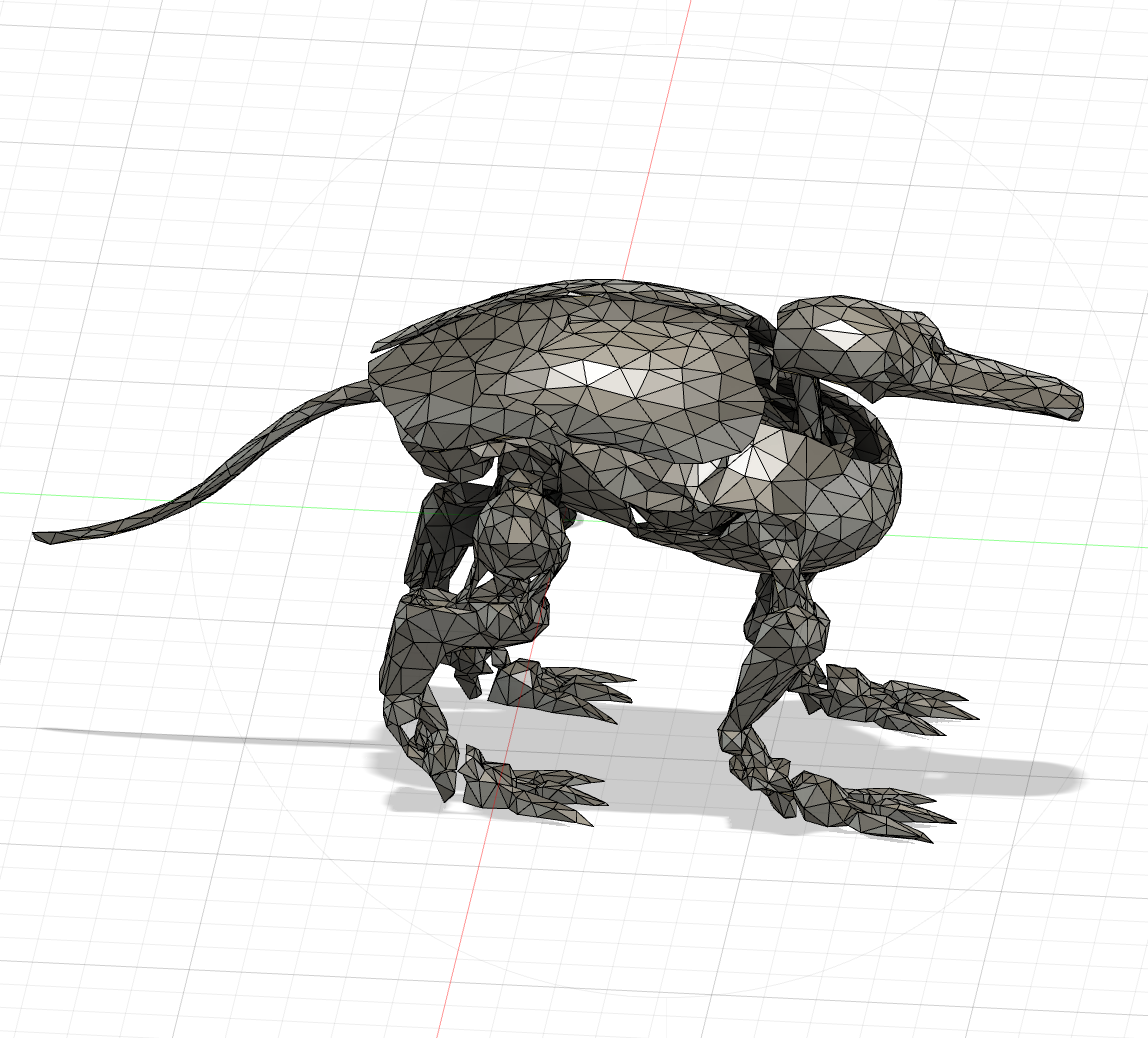} &
    \includegraphics[width=0.122\textwidth]{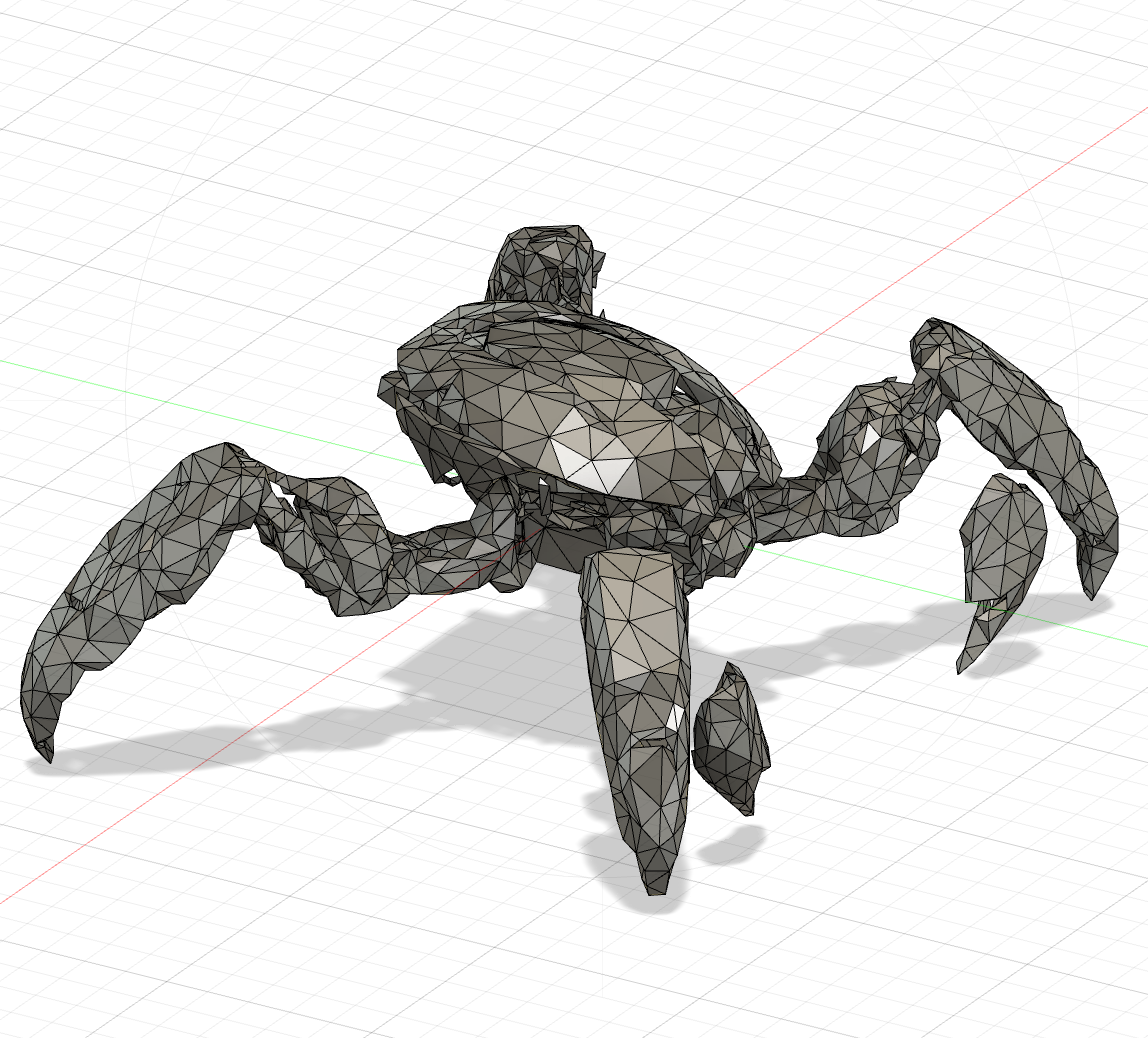} \\

    Seahorse & Kingfisher & Platypus & Crab

    \end{tabular}
    
    \caption{Comparison of  designs generated by our method (top row) vs Text2Robot~\cite{ringel2024text2robot} (bottom row)}
    \label{fig:baseline_comparison}	
\end{figure}

Text2Robot (T2R) leverages Meshy~\cite{meshy} for mesh initialization, followed by a geometric and evolutionary pipeline for quadruped robots. While T2R excels at rapidly creating physical robots whose initial structures match text-specified aesthetics, its morphological expressiveness is ultimately bounded by the capabilities of the generator and post-processor, focusing primarily on quadrupeds with fixed-limb and joint templates. In contrast, our framework is both complementary and strictly more general at the asset generation stage. Because our pipeline synthesizes robot kinematics and structure natively, it can realize arbitrary morphologies. 
We provide a comparison to non-quadrupedal creatures in Figure~\ref{fig:baseline_comparison}, where T2R generates four legs for all Seahorse, Kingfisher, and Crab examples. 
We believe our method can be synergistic with T2R by providing better initialization and structural understanding.

\section{LIMITATIONS}

Although \textbf{RobotDesignGPT} can synthesize robot models that appear realistic, there is still significant room for improvement. Our framework encounters several challenges due to the reliance on Vision-Language Models (VLMs) which were not specifically trained using robotics data. These include spatial reasoning issues, hallucinated symmetry, coordinate frame confusion, stochastic behaviors producing different kinematic trees for the same morphology, and occasional gaps between components. Certain delicate structures, such as wings and fins, are particularly challenging to model. These structures are often incorrectly generated as large spheres or blocks. The quality of the visual feedback results shows a significant dependence on the choice of reference images.


Joint-related issues are commonly observed, including incorrect bidirectional actuation for joints like knees, inappropriate joint type selection (such as ball joints instead of fixed joints), and unrealistic joint axis configurations. While these issues can be addressed through human feedback, they represent a notable limitation of the current system.

Finally, our design framework focuses primarily on kinematic structures and visual appearance. We exclude the synthesis of dynamic properties, which is beyond the scope of this work. These properties may include parameters such as PD gains, maximum joint torques, and inertial properties. While VLMs can provide reasonable initial guesses for these properties, incorporating feedback from trajectory optimization or policy learning frameworks remains an interesting direction for future work.

\section{CONCLUSIONS}

We introduced \textbf{RobotDesignGPT}, a novel framework for robot design synthesis that leverages the strengths of VLMs. Our system enables users to generate kinematically valid and visually realistic robot designs by simply providing textual descriptions and reference images. Through iterative visual and human feedback mechanisms, the framework refines designs to ensure visual fidelity and kinematic coherency. We demonstrate that our framework can synthesize a diverse range of robots, such as bee, kingfisher, and bison, with an average of $2.15$ iterations of human feedback. The generated designs are not only visually accurate, achieving an average user rating of $2.93$ out of five for accuracy, but also kinematically valid, as confirmed by successful integration with motion planning modules. In the future, we hope to investigate a broader range of extensions, such as dynamic properties or task-based robot design. A fine-tuning of the existing LLMs and VLMs for robot design synthesis would also be an interesting future research direction.







\bibliographystyle{IEEEtran}
\bibliography{refs}

\end{document}